%% file: main.tex
\documentclass{article}

\usepackage{iclr2027_conference,times}

\usepackage{amsmath,amsfonts}

\usepackage{graphicx}
\graphicspath{{figures/}}
\usepackage{hyperref}
\hypersetup{
  hidelinks,
  pdftitle={UniAction: Unified Action Conditioning for World Models},
  pdfauthor={Bowen Xue, Zizhang Li, Hong-Xing Yu, Jiajun Wu}
}
\usepackage{xspace}
\usepackage{enumitem}
\usepackage{booktabs}
\usepackage{multirow}
\usepackage{array,colortbl}
\usepackage{wrapfig}
\usepackage{needspace}

\newcommand{\ourmethod}{UniAction\xspace}
\renewcommand{\paragraph}[1]{\vspace{0.1cm}\noindent\textbf{#1}}
\title{UniAction: Unified Action Conditioning for\\World Models}

\author{Bowen Xue\thanks{Equal contribution.}\\
\normalfont Stanford University
\And
Zizhang Li\footnotemark[1]\\
\normalfont Stanford University
\And
Hong-Xing Yu\\
\normalfont Stanford University
\And
Jiajun Wu\\
\normalfont Stanford University}

\iclrfinalcopy

\begin{document}

\maketitle
\fancyhead{}
\renewcommand{\headrulewidth}{0pt}

\begin{abstract}
  \input{sections/00_abstract}
\end{abstract}

\input{sections/01_introduction}

\input{sections/02_related_work}
\input{sections/03_method}
\input{sections/03_instantiation}
\input{sections/04_experiments}
\input{sections/05_conclusion}

\newpage
\input{sections/06_ai_use_statement}
\input{sections/07_ethics_statement}

\bibliography{references}
\bibliographystyle{iclr2027_conference}

\newpage
\appendix
\input{appendices/appendix}

\end{document}

%% file: sections/00_abstract.tex
Actions that condition world models are highly heterogeneous, and thus existing action-conditioned world models are typically built for individual action types, fragmenting already limited action-annotated data across separate models.
We introduce UniAction, a unified action-conditioning framework that represents any action as a sparse set of action elements, each with a discrete semantic identity and a continuous spatiotemporal grounding coordinate.
By mapping these coordinates directly into the pretrained video model's native positional space, UniAction allows heterogeneous actions to condition a shared world model within the same spatiotemporal structure. We further introduce cardinality-calibrated attention to address the bias from differing action representation granularities. We evaluate UniAction on four representative action forms: force, camera control, hand pose, and robot pose. UniAction outperforms conditioning alternatives in action following, and a single jointly trained model surpasses the action-specific experts, showing that heterogeneous action data benefit from unified model and joint training.
Project page: \textbf{\url{https://uniaction.github.io/}}

%% file: sections/01_introduction.tex
\section{Introduction}

A world model predicts how the world evolves in response to actions~\citep{parker2025genie,team2026advancing}, and large-scale video generation models~\citep{minimax2026h3, wan2025wan}, which learn rich spatiotemporal dynamics from real-world videos, provide a natural foundation for building the world models~\citep{brooks2024video}. The actions that condition such world models, however, are highly heterogeneous. They range from global viewpoint changes~\citep{parker2025genie,li2025hunyuan,tang2025hunyuan,sun2025worldplay,he2025matrix} and physical forces applied to objects~\citep{gillman2025force,gillman2026goal,physctrl2025,wang2026streaming} to hand and robot motions~~\citep{sudhakar2024controlling,xie2026generated,wang2025precise,zhu2025irasim}, and differ greatly in their semantics and spatiotemporal structures.

Given this heterogeneity, most existing action-conditioned world models are built for a single action type, with a conditioning mechanism designed for that type and training restricted to its data~\citep{li2025hunyuan,sun2025worldplay,he2025matrix,gillman2025force,wang2026streaming,sudhakar2024controlling,zhu2025irasim}.
This practice is costly because action-annotated data is scarce: compared with large-scale generic video, it is limited in scale and concentrated in a narrow range of scenes and interaction patterns~\citep{gillman2025force}. Training separate models further fragments this data, and since each model learns how the world responds to only one type of action, knowledge of scene structure, object properties, and world dynamics that could be shared remains isolated across models~(Fig.~\ref{fig:teaser}, left).
As the number of action types grows, such separate training scales poorly.

To bring data from different actions into a single world model, we first need a conditioning representation that can accommodate their heterogeneous forms in a unified way. Existing generic mechanisms offer two natural choices. One renders actions into image or video conditions~\citep{wang2025precise,wu2026oscar}, providing direct spatial correspondence with the generated content, but not every action has a visual form that can be directly rendered. The other encodes different actions into conditioning tokens~\citep{li2025multimodal,nvidia2025cosmos}, which flexibly accommodates signals of different forms, but where and when an action takes effect in the generated video remains an implicit correspondence. These limitations lead us to reconsider what information heterogeneous actions actually need to convey to a world model.

Our key observation is that, despite their heterogeneous raw forms, every action conveys two complementary pieces of information to the world model: what the action is, and where and when it takes effect.
Based on this observation, we introduce \ourmethod, which represents each action as a sparse set of action elements. Each element describes its semantics with a discrete identity and its spatiotemporal grounding with a continuous coordinate. We place these coordinates directly in the pretrained video model's native positional geometry, allowing action and video to interact within a shared spatiotemporal structure while reusing the model's pretrained spatiotemporal modeling capabilities. Heterogeneous actions can thus condition a single model through the same mechanism without requiring a common raw representation or rendering into visual conditions. Since different actions may be represented by different numbers of elements, we further introduce cardinality-calibrated attention to prevent representation granularity from biasing conditioning strength.

We instantiate \ourmethod on four representative action forms: force, camera control, hand pose, and robot pose. Under matched training settings, \ourmethod outperforms alternative conditioning mechanisms in both action following and visual quality. More importantly, a single jointly trained \ourmethod model consistently outperforms the corresponding action-specific experts across all four action forms~(Fig.~\ref{fig:teaser}, right), demonstrating the effectiveness of unified action conditioning and the benefit of jointly learning from heterogeneous action data. Our contributions are threefold:

\begin{itemize}[nosep,leftmargin=*]
\item We introduce \ourmethod, a unified action-conditioning framework that represents heterogeneous actions through action semantics and spatiotemporal grounding, placing action coordinates in the pretrained video model's native positional geometry so that all actions condition the model through the same mechanism.

\item We introduce cardinality-calibrated attention to mitigate the bias caused by varying numbers of action tokens, reducing the influence of representation granularity on conditioning strength.

\item We evaluate \ourmethod across four representative action forms. \ourmethod achieves better action following and visual quality than visual- and token-based conditioning alternatives, while a single jointly trained model outperforms the corresponding action-specific experts, showing the benefit of jointly learning from heterogeneous action data.
\end{itemize}

\input{figTexts/fig_teaser}

%% file: figTexts/fig_teaser.tex
\begin{figure}[t]
  \centering
   \includegraphics[width=\linewidth]{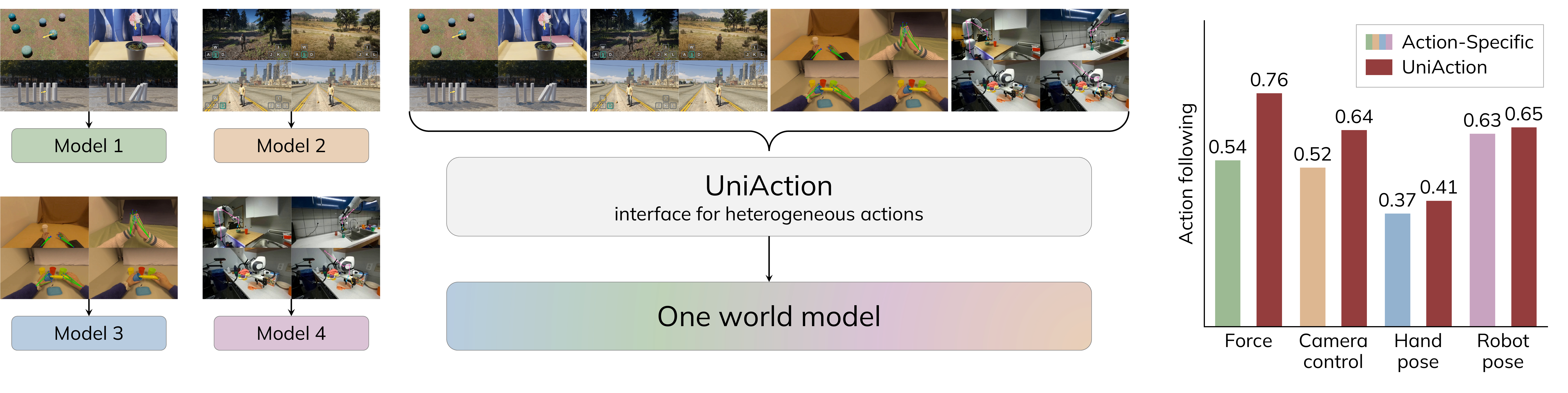}

   \caption{\textbf{One world model for heterogeneous actions.} Instead of training a separate model per action type~(left), \ourmethod conditions a single world model~(middle) on heterogeneous actions and outperforms the action-specific experts across all four action forms~(right).}
\label{fig:teaser}
\end{figure}

%% file: sections/02_related_work.tex
\section{Related Work}

\paragraph{Video World Models.}
Large-scale video generation models, such as Wan~\citep{wan2025wan} and MiniMax~\citep{minimax2026h3}, learn visual structures and spatiotemporal dynamics through video pretraining, providing a generative foundation for predicting how the world evolves from current observations.
Building on these powerful video models, a line of works~\citep{parker2025genie,li2025hunyuan,he2025matrix,sun2025worldplay,gao2026infinite} further
introduce user interactions, enabling the models to continuously generate future observations of an environment under the given controls. These video world models typically define their action spaces and control interfaces for a specific environment; for example, the keyboard inputs and camera movement signals~\citep{li2025hunyuan,he2025matrix}. In contrast, \ourmethod builds on the same class of pretrained video models and allows actions from different environments with different structures to condition a single video world model through a shared interface.

\paragraph{Action Conditioning in Video Models.}
Existing action-conditioned video models largely focus on specific action spaces. One line of work studies game commands, camera motion, and navigation controls~\citep{li2025hunyuan,he2025matrix,sun2025worldplay,team2026advancing,gao2026infinite},
while another models force-based physical interventions~\citep{gillman2025force,gillman2026goal,physctrl2025,wang2026streaming} with the environment.
Other approaches condition video models on hand motion~\citep{sudhakar2024controlling} or robot manipulation trajectories~\citep{zhu2025irasim}. These methods typically use representations and conditioning mechanisms tailored to their respective action spaces.

Recent work has begun to explore unifying related action spaces. Shared skeletal or kinematic representations connect human and robot motion~\citep{wang2025precise,wu2026oscar}, while the recent Cosmos 3~\citep{agarwal2026cosmos} represents controls across embodied systems through common physical variables.
These approaches build on shared kinematic structures or common physical-control parameterizations. In contrast, \ourmethod accommodates heterogeneous actions with distinct structures and parameterizations through a shared formulation of action semantics and spatiotemporal grounding, enabling joint learning from diverse action-conditioned data within a single world model.

\paragraph{Controllable Video Generation.}
Controllable video generation typically achieves unified control by converting external signals into shared conditioning representations
compatible with the video model.
Several approaches represent controls as visual conditions aligned with the generated video~\citep{wang2023videocomposer,zhang2024controlvideo,jiang2025vace}. However, this requires a suitable visual encoding for each control signal, which is less natural for abstract interventions such as forces. Alternatively, separate learned encoders can map different sensory inputs and actions into features that are fused and injected through a cross-attention mechanism~\citep{li2025multimodal}. These approaches accommodate heterogeneous inputs through either visual renderings or learned encoders for specific action formats. In contrast, \ourmethod directly represents actions following the same conditioning formulation at the representation level, without requiring a separate renderer or learned encoder for each action type.

%% file: sections/03_method.tex
\section{Method}

\input{figTexts/fig_method}

Given an initial frame $I_0$, a text prompt $T$, and an action $A$, we model the conditional generation of a future video $V$ as
\begin{equation}
    p_\theta(V \mid I_0, T, A).
\end{equation}
UniAction represents any action as a sparse set of action elements:
\begin{equation}
    A = \left\{(s_i,\mathbf{p}_i)\right\}_{i=1}^{N},
    \qquad
    \mathbf{p}_i = (t_i,y_i,x_i),
\end{equation}
where $s_i$ denotes the semantic identity of the $i$-th element and $\mathbf{p}_i$ its continuous grounding coordinate in video space-time. Different actions may use different semantic identities and numbers of elements, but all follow this unified representation and condition a shared video model through the same mechanism.

\subsection{Action Semantics}
\label{sec:action_semantics}

Action semantics describe the meaning represented by each action element. We define a shared discrete semantic vocabulary $\mathcal{S}$ that distinguishes semantic identities both within and across action forms. A semantic codebook
$E \in \mathbb{R}^{|\mathcal{S}| \times d}$
associates each identity with a $d$-dimensional action token, where $d$ matches the feature dimension of the video tokens. For the $i$-th action element, its semantic identity $s_i \in \mathcal{S}$ directly indexes the corresponding action token: $\mathbf{a}_i = E[s_i] \in \mathbb{R}^{d}$.

Each semantic identity is associated with an independently initialized action token, and the model learns its correspondence to scene evolution through joint action-conditioned training. Elements sharing the same identity reuse the same action token across different locations and times, while these spatiotemporal differences are modeled separately by the spatiotemporal grounding described in the next section. New semantic identities can be added directly to the same vocabulary, allowing the representation to naturally extend to new action forms.

\subsection{Spatiotemporal Grounding}
\label{sec:spatiotemporal_grounding}

Spatiotemporal grounding specifies where and when an action element takes effect. Although heterogeneous actions may be defined in different coordinate and control spaces, their effects ultimately manifest in the spatiotemporal evolution of the generated video. We therefore use the video as a common spatiotemporal reference and represent the grounding of the $i$-th action element with a continuous coordinate $\mathbf{p}_i=(t_i,y_i,x_i)$.

To align action grounding with the spatiotemporal structure established by the pretrained video model, UniAction directly reuses the backbone's native positional system. Let $\Phi_{\mathcal B}(\mathbf{x},\mathbf{p})$ denote the backbone's positional mechanism for a token $\mathbf{x}$ at coordinate $\mathbf{p}$. For an action token $\mathbf{a}_i$ and a video token $\mathbf{v}_j$,
\begin{equation}
\widetilde{\mathbf{a}}_i
= \Phi_{\mathcal B}(\mathbf{a}_i,\mathbf{p}_i),
\qquad
\widetilde{\mathbf{v}}_j
= \Phi_{\mathcal B}(\mathbf{v}_j,\mathbf{p}^{v}_j),
\end{equation}
where $\mathbf{p}^{v}_j$ denotes the spatiotemporal coordinate of the video token. Action and video tokens thus establish their spatiotemporal relations within the same positional geometry and interact through the backbone's native self-attention~\citep{vaswani2017attention}. Since this positional geometry is already established during video pretraining, action--video interactions can directly reuse the backbone's pretrained spatiotemporal modeling capabilities.

\subsection{Cardinality-Calibrated Attention}
\label{sec:cardinality_calibration}

Different action forms can be represented at different granularities, leading to potentially large differences in the number of action tokens. Since action and video tokens are jointly processed by the backbone's self-attention, we examine how token cardinality affects attention allocation. For a video query, let $\ell_i^a$ denote its attention logit to the $i$-th action token. For an action represented by $N$ tokens, its total unnormalized attention mass is
\begin{equation}
Z_a
= \sum_{i=1}^{N}\exp(\ell_i^a)
= N\cdot\operatorname{mean}_i[\exp(\ell_i^a)].
\end{equation}

The factor $N$ is determined solely by representation granularity and is independent of the average action--video compatibility, thereby introducing an additional cardinality bias. To remove this bias, we apply the same calibration term to all action logits:
\begin{equation}
\hat{\ell}_i^a
= \ell_i^a+\log\frac{N_{\mathrm{ref}}}{N},
\end{equation}
where $N_{\mathrm{ref}}$ is a fixed reference cardinality shared across all actions, while the video logits remain unchanged.
After calibration, the total action mass becomes
$N_{\mathrm{ref}}\cdot\operatorname{mean}_i[\exp(\ell_i^a)]$,
removing its dependence on the actual number of action tokens and placing different actions on a common attention scale. Since the same offset is added to all action logits, the relative attention relationships among action tokens remain unchanged. We detail the mathematical proof in Supplementary~\ref{sec:supp_cardinality}.

%% file: figTexts/fig_method.tex
\begin{figure}[t]
  \centering
   \includegraphics[width=\linewidth]{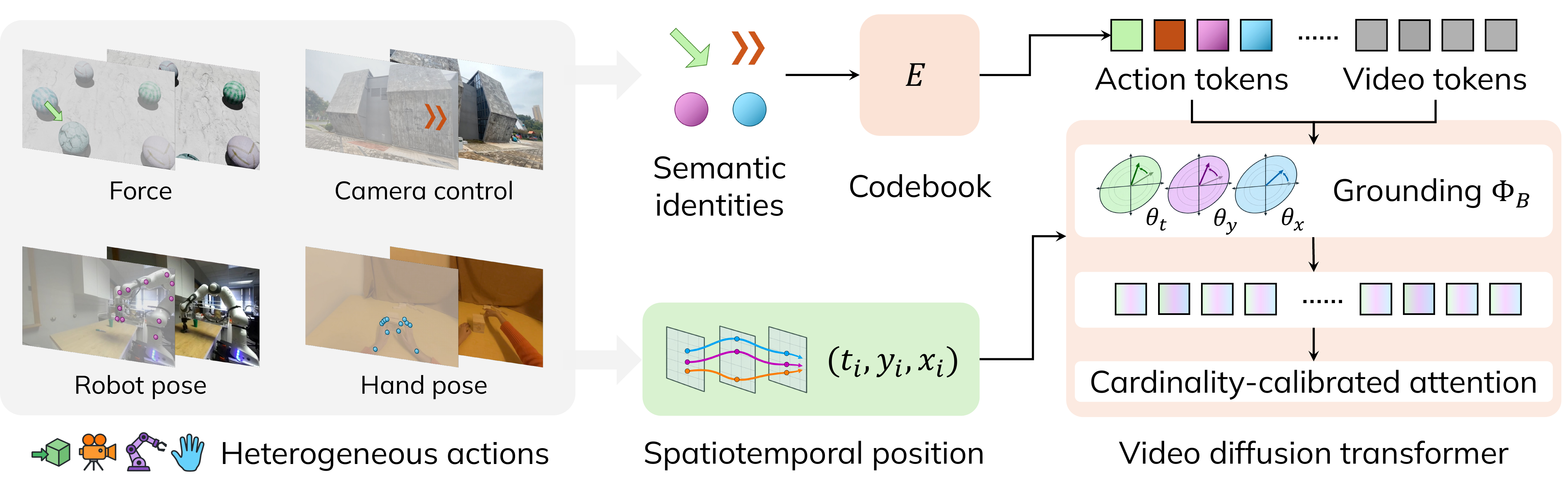}

   \caption{\textbf{Overview of \ourmethod.} Diverse action forms are represented as sparse action elements: semantic identities index tokens in a shared codebook~(Sec.~\ref{sec:action_semantics}), spatiotemporal positions use the backbone's native positional encoding~(Sec.~\ref{sec:spatiotemporal_grounding}), and action and video tokens interact via cardinality-calibrated attention~(Sec.~\ref{sec:cardinality_calibration}).}
\end{figure}

%% file: sections/03_instantiation.tex
\section{Experiments}

We first describe how \ourmethod is instantiated in a pretrained video backbone~(Sec.~\ref{sec:backbone_instantiation}) and on four representative diverse action forms~(Sec.~\ref{sec:action_instantiation}), and then present the experimental setup~(Sec.~\ref{sec:exp_setup}), and the results on comparisons and ablations in the following sections.

\subsection{Backbone instantiation}

\label{sec:backbone_instantiation}
Our model backbone, Wan2.2~\citep{wan2025wan}, is a latent video transformer with 3D RoPE, into which UniAction is instantiated directly through its existing token sequence and self-attention.
The semantic identity $s_i$ of each action element indexes an action token $\mathbf{a}_i$ as in Sec.~\ref{sec:action_semantics}, and all action tokens are concatenated after the latent video token sequence:
    $[\,\mathbf{v}_1,\ldots,\mathbf{v}_M,\ \mathbf{a}_1,\ldots,\mathbf{a}_N\,].$

The action location $\mathbf{p}_i$ in video space is mapped into the backbone's latent coordinate system through its own VAE~\citep{kingma2014vae} compression and tokenization,
\begin{equation}
    \mathbf{p}_i^{\mathrm{lat}}
    =
    g_{\mathcal B}
    \left(
        \mathbf{p}_i
    \right),
\end{equation}
and in self-attention the same rotary positional encoding~\citep{su2024roformer} is applied to action tokens at $\mathbf{p}_i^{\mathrm{lat}}$ and to video tokens at their latent grid coordinates $\mathbf{p}^{\mathrm{v}}_j$.

Consequently, for two action elements that share a semantic identity but occur at different locations,
\begin{equation}
    s_i = s_j
    \implies
    \mathbf{a}_i = \mathbf{a}_j = E[s_i],
    \qquad
    \mathbf{p}_i^{\mathrm{lat}} \neq \mathbf{p}_j^{\mathrm{lat}},
\end{equation}
so a semantic identity can be reused across locations, while where it takes effect is distinguished entirely by the backbone's native positional geometry.

\subsection{Action instantiation}

\label{sec:action_instantiation}
The four action types, force, camera control, hand pose, and robot pose, differ substantially in their raw forms:
some are instantaneous physical vectors, some are discrete commands with no intrinsic spatial location, and some are dense per-frame, per-joint annotations.

\paragraph{Hand pose.} A hand pose is a dense per-frame, per-joint action, and the image location of each joint is itself where the intervention takes effect. The semantic identities are joint indices, one per joint; at each frame, every visible joint yields one action element located at that joint's position in the frame.

\paragraph{Robot pose.} A robot pose is given by a mechanical structure of rigid links, whose number of joints and geometry are determined by the particular robot. The semantic identities are robot joint indices; at each frame, every visible joint yields one action element at that joint's position in the frame.

\paragraph{Camera control.}
We consider keyboard-based camera control, where viewpoint changes are specified by discrete key commands. The semantic identities correspond to eight camera commands: four translations~(forward, backward, left, right; keys W, S, A, D) and four rotations~(pitch up, pitch down, yaw left, yaw right; keys I, K, J, L). Since these commands act on the global view rather than at a specific location, they have no intrinsic spatial grounding. We therefore assign each direction a fixed canonical location on the corresponding side of the frame; translation and rotation commands with the same direction share this location and are distinguished by their semantic identities. At each frame, every active command yields one action element at its canonical location and the current time.

\paragraph{Force.} A force is an instantaneous vector applied at a point, and its direction and magnitude cannot be expressed by a single location. We represent one applied force with two elements at the same instant, one at the tail and one at the head of the force vector, with the semantic identities marking these two roles. The spatial offset between them is the force vector.

%% file: sections/04_experiments.tex
\setlength{\textfloatsep}{14pt plus 2pt minus 2pt}
\subsection{Experimental Setup}
\label{sec:exp_setup}

\input{tables/comparison}
\input{figTexts/fig_qualitative}
\input{figTexts/fig_expert_comparison}

We summarize the data, implementation, and evaluation protocol below; full configurations and details are given in Supplementary~\ref{sec:supp_experimental_details}.

\Needspace{5\baselineskip}
\paragraph{Datasets.}
The training data for force, camera control, hand pose, and robot pose are drawn from
GoalForce~\citep{gillman2026goal}, SolarWM~\citep{huang2026solarwm}, EgoDex~\citep{hoque2026egodex}, and DROID~\citep{khazatsky2024droid}, respectively. We sample 10K clips from each dataset, 40K in total, each with 81 frames at $480\times832$ resolution.

\paragraph{Implementation details.}
Our main experiments use the high-noise expert of Wan2.2-I2V-A14B~\citep{wan2025wan} as the generative backbone and adapt it with LoRA~\citep{hu2022lora}.

\paragraph{Evaluation.}
We primarily evaluate \textit{Action Following}, i.e., whether the generated video evolves as specified by the given action. For each test sample, we fix the initial frame and text prompt while varying only the action condition, such as rotated or reversed forces, different camera directions, and flipped or stationary pose trajectories. This yields 410 counterfactual generations that isolate the model's response to the specified intervention.
We then measure each response with an off-the-shelf estimator: DIS optical flow~\citep{kroeger2016dis} to estimate object motion in force scenes, CoTracker3~\citep{karaev2025cotracker3} to track hand keypoints for hand pose and control points for robot pose, and ViPE~\citep{huang2025vipe} to estimate camera pose trajectories for camera control. From these motion measurements, we compute an Action Following score for each action form and normalize all scores to $[0,1]$, where higher values indicate a more accurate response to the given action. Full metric definitions and normalization are provided in the supplementary material. We additionally report LPIPS~\citep{zhang2018lpips} on original condition's generation as a complementary measure of perceptual similarity to the ground truth frames.

In addition, we conduct a human evaluation on 60 cases sampled uniformly across the four action forms, with 100 participants. In each case, the videos generated by the three methods are presented together, with method names hidden and their order randomized. Participants rate each video on a 1--5 scale along two dimensions: \emph{Action Following}, i.e., whether the video performs the given action, and \emph{Visual Quality}, covering clarity, stability, and the plausibility of physical interactions. We report the average scores; the full evaluation protocol is provided in Supplementary~\ref{sec:supp_experimental_details}.

\subsection{Comparison with Generic Conditioning Mechanisms}

We compare \ourmethod with two generic conditioning mechanisms.
All three methods start from the same pretrained model and are trained jointly on the same 40K samples of four action types with matched optimization settings.
They differ only in how actions are represented and injected into the video model, so the comparison isolates the conditioning mechanism itself.

\paragraph{Visual Conditioning~(VACE).}
We adopt a VACE~\citep{jiang2025vace} style visual conditioning baseline that renders different actions as control videos, encodes them with the VAE, and feeds them into the VACE conditioning branch. Force is represented by an application point and a directional arrow, camera control by temporally highlighted key indicators, and hand and robot poses by their corresponding skeletons. The VACE conditioning branch consists of eight replicated video DiT blocks,
with approximately $3.1$B trainable parameters.

\paragraph{Token Conditioning~(Cross-Attn).}
We adopt a cross-attention conditioning baseline~\citep{nvidia2026worldsimulationvideofoundation}. We use separate encoders for different action forms, to directly map action signals and their spatiotemporal information into conditioning tokens of a shared dimensionality, which are then injected into the video model through cross-attention layer. The action encoders and cross-attention modules are both trained, with LoRA applied to
the cross-attention modules, resulting in approximately $640$M trainable parameters.

\paragraph{UniAction.}
All action forms share the same conditioning pathway, where action tokens interact with video tokens through the backbone's native self-attention. \ourmethod{} trains a shared action-branch LoRA that is applied only to action tokens, with approximately 404M trainable parameters.

As shown in Table~\ref{tab:conditioning_comparison}, \ourmethod achieves the best action following performance across all four action forms, together with the lowest LPIPS.
Human evaluation leads to a consistent conclusion: UniAction receives the highest action following ratings on all action forms, along with comparable or better overall visual quality.
With the spatiotemporal information in a learned encoder space rather than the backbone's positional geometry, Cross-Attn performs markedly worse on actions that require spatial localization like robot and hand pose.
VACE is competitive where an action has a natural visual rendering, but with the largest gap to \ourmethod on force and camera control, which lack a native visual form. In contrast, \ourmethod responds effectively to both types of actions through semantic identities and explicit spatiotemporal coordinates. Figure~\ref{fig:qualitative_comparison} presents qualitative comparisons across the four action forms, more results in Supplementary~\ref{sec:supp_qualitative}.

\vspace{-5pt}
\subsection{Benefits of Unified Training}
We next examine whether heterogeneous action data benefit one another when trained within a single model. To test this, we train an action-specific expert baseline for each action form. Each expert has the same architecture as \ourmethod{} and is trained on the data of its corresponding action form. For each action form, the expert and the jointly trained \ourmethod{} use exactly the same training samples, and each sample is seen the same number of times. The four experts are thus equivalent to a single model with $4\times$ the parameters that hard-routes samples by action form, trained with the same total compute as the jointly trained \ourmethod.

\input{tables/joint_expert}As shown in Table~\ref{tab:joint_learning}, the jointly trained \ourmethod outperforms the corresponding experts in action following across all four action forms. The jointly trained model also achieves lower or comparable LPIPS. Figure~\ref{fig:expert_comparison} further shows the qualitative comparisons.

We attribute these gains to shared learning of world dynamics. An expert learns how the world responds to actions only from the limited annotated data of one action form. In contrast, the jointly trained model learns scene properties and dynamics from all four forms, so each form's data is complemented by data from the others.

\subsection{Ablation Studies}

We ablate the two key designs of \ourmethod: spatiotemporal grounding and cardinality calibration. All ablation studies use the Wan2.2-TI2V-5B~\citep{wan2025wan} model under the same training and evaluation protocol. Quantitative results are reported in Table~\ref{tab:ablation}.

\subsubsection{Spatiotemporal Grounding}
\label{sec:ablation_grounding}

\ourmethod maps action coordinates into the backbone's native positional system, so action and video tokens share the same positional geometry learned during video pretraining. We test this design with three variants, each changing one aspect:

\textbf{Self-attn $\rightarrow$ Cross-attn}
keeps the RoPE~\citep{su2024roformer} encoding of action coordinates but injects action tokens through the text embedding's cross-attention pathway instead of self-attention. In this case, action and video tokens' coordinates no longer share the positional system. This variant tests whether encoding action coordinates is sufficient on its own, without the shared positional geometry.

\textbf{RoPE $\rightarrow$ Sinusoidal APE} and \textbf{RoPE $\rightarrow$ Learned APE}
keep the shared self-attention but encode action coordinates with, respectively, fixed sinusoidal absolute positional \mbox{encodings} \mbox{\citep{vaswani2017attention}} and learned absolute positional embeddings~\citep{dosovitskiy2021image}, instead of the \mbox{backbone's} pretrained 3D RoPE. These variants test whether reusing the pretrained positional geometry matters, as opposed to merely sharing an attention pathway.

\input{tables/ablation}\looseness=-1
As shown in Table~\ref{tab:ablation},
under the same training budget, all three variants achieve low action following across all action forms and nearly fail on hand pose and robot pose, which require precise spatial grounding. These variants all need to learn the spatiotemporal correspondence between actions and videos anew, whereas UniAction places action coordinates in the backbone's pretrained positional system and directly reuses its existing spatiotemporal modeling capabilities, thereby acquiring precise control from limited action data.

\input{figTexts/fig_temporal_generalization}\paragraph{Temporal Generalization.}
UniAction reuses the backbone's pretrained positional geometry to encode action coordinates, enabling generalization to unseen action times. In the force training data, forces are always applied at the first frame~($t=0$). At test time, we apply forces at $t=8$ and $t=24$~(Figure~\ref{fig:temporal_generalization}). \ourmethod starts the induced motion at the specified time, whereas Learned APE changes little as the force time shifts because unseen action-time embeddings receive no supervision. This suggests that UniAction leverages temporal structure learned during video pretraining rather than relearning action-time positions from limited action data.

\vspace{-4pt}
\subsubsection{Cardinality Calibration}
\vspace{-3pt}

To evaluate the cardinality calibration introduced in Sec.~\ref{sec:cardinality_calibration}, we remove the calibration term during both training and inference and keep everything else unchanged.
As shown in Table~\ref{tab:ablation}, removing calibration leads to clear drops in action following for force and camera control, whereas pose actions remain largely unaffected.

This pattern follows the analysis in Sec.~\ref{sec:cardinality_calibration}.
Without calibration, the total attention mass grows with the number of tokens: pose actions, represented by many tokens, naturally dominate the attention, while force and camera control, represented by only a few tokens, are suppressed. Cardinality calibration removes this bias and places different actions on a common attention scale, allowing actions with sparse tokens to control generation just as effectively.

%% file: tables/comparison.tex
\begin{table}[t]
\centering
\vspace{-3mm}
\caption{Comparison with visual-based~(VACE~\citep{jiang2025vace}) and token-based (Cross-Attn~\citep{nvidia2026worldsimulationvideofoundation}) conditioning mechanisms across four action forms. Best and second-best results are \textbf{bold} and \underline{underlined}.}
\label{tab:conditioning_comparison}
\begingroup
\fontsize{9}{11}\selectfont
\definecolor{uniactionrow}{gray}{0.92}
\setlength{\tabcolsep}{0pt}
\renewcommand{\arraystretch}{1.22}
\arrayrulecolor{black}
\newcommand{\metriccell}[1]{\makebox[\dimexpr(\linewidth-152pt)/4\relax][c]{#1}}
\newcommand{\metricheading}[2]{\metriccell{\(\vcenter{\hbox{\shortstack[c]{#1}}}\,#2\)}}
\newcommand{\methodcell}[1]{\makebox[72pt][c]{#1}}
\newcommand{\tableheading}[2]{\multirow{2}{#1}[-3.7pt]{\centering\textbf{#2}}}
\newcommand{\actioncell}[1]{\multirow{-3}{74pt}{\centering\textbf{\shortstack[c]{#1}}}}
\newcommand{\ourscell}[1]{\cellcolor{uniactionrow}#1}
\newcommand{\actionrule}{\arrayrulecolor{black!35}\midrule[0.25pt]\arrayrulecolor{black}}
\begin{tabular}{@{}l@{\hspace{6pt}}lcccc@{}}
\toprule[0.8pt]
\tableheading{74pt}{Action Form}
& \tableheading{72pt}{Method}
& \multicolumn{2}{c}{\textbf{Automatic Evaluation}}
& \multicolumn{2}{c}{\textbf{Human Evaluation}} \\
\cmidrule(lr){3-4}\cmidrule(l){5-6}
&
& \metricheading{Action\\Following}{\uparrow} & \metricheading{LPIPS}{\downarrow} & \metricheading{Action\\Following}{\uparrow} & \metricheading{Visual\\Quality}{\uparrow} \\
\midrule[0.4pt]
& \methodcell{VACE} & \underline{0.6046} & \underline{0.1178} & \underline{2.47} & \underline{3.40} \\
& \methodcell{Cross Attention} & 0.1618 & 0.1341 & 1.96 & 3.13 \\
\actioncell{Force}
& \ourscell{\methodcell{\textbf{UniAction} (Ours)}} & \ourscell{\textbf{0.7630}} & \ourscell{\textbf{0.1163}} & \ourscell{\textbf{4.37}} & \ourscell{\textbf{4.00}} \\
\actionrule
& \methodcell{VACE} & 0.4997 & \underline{0.5100} & \underline{3.30} & \underline{3.55} \\
& \methodcell{Cross Attention} & \underline{0.5306} & 0.5124 & 3.13 & 3.35 \\
\actioncell{Camera Control}
& \ourscell{\methodcell{\textbf{UniAction} (Ours)}} & \ourscell{\textbf{0.6371}} & \ourscell{\textbf{0.4863}} & \ourscell{\textbf{4.02}} & \ourscell{\textbf{3.68}} \\
\actionrule
& \methodcell{VACE} & \underline{0.2938} & \underline{0.1122} & \underline{3.30} & \textbf{3.43} \\
& \methodcell{Cross Attention} & 0.0678 & 0.1695 & 2.34 & 3.12 \\
\actioncell{Hand Pose}
& \ourscell{\methodcell{\textbf{UniAction} (Ours)}} & \ourscell{\textbf{0.4119}} & \ourscell{\textbf{0.0910}} & \ourscell{\textbf{3.36}} & \ourscell{\underline{3.42}} \\
\actionrule
& \methodcell{VACE} & \underline{0.5245} & \underline{0.1389} & \underline{3.60} & \underline{3.73} \\
& \methodcell{Cross Attention} & 0.0859 & 0.1859 & 2.07 & 3.44 \\
\actioncell{Robot Pose}
& \ourscell{\methodcell{\textbf{UniAction} (Ours)}} & \ourscell{\textbf{0.6521}} & \ourscell{\textbf{0.1164}} & \ourscell{\textbf{4.32}} & \ourscell{\textbf{3.99}} \\
\bottomrule[0.8pt]
\end{tabular}
\endgroup
\end{table}

%% file: figTexts/fig_qualitative.tex
\begin{figure}[t]
  \centering
  \includegraphics[width=\linewidth]{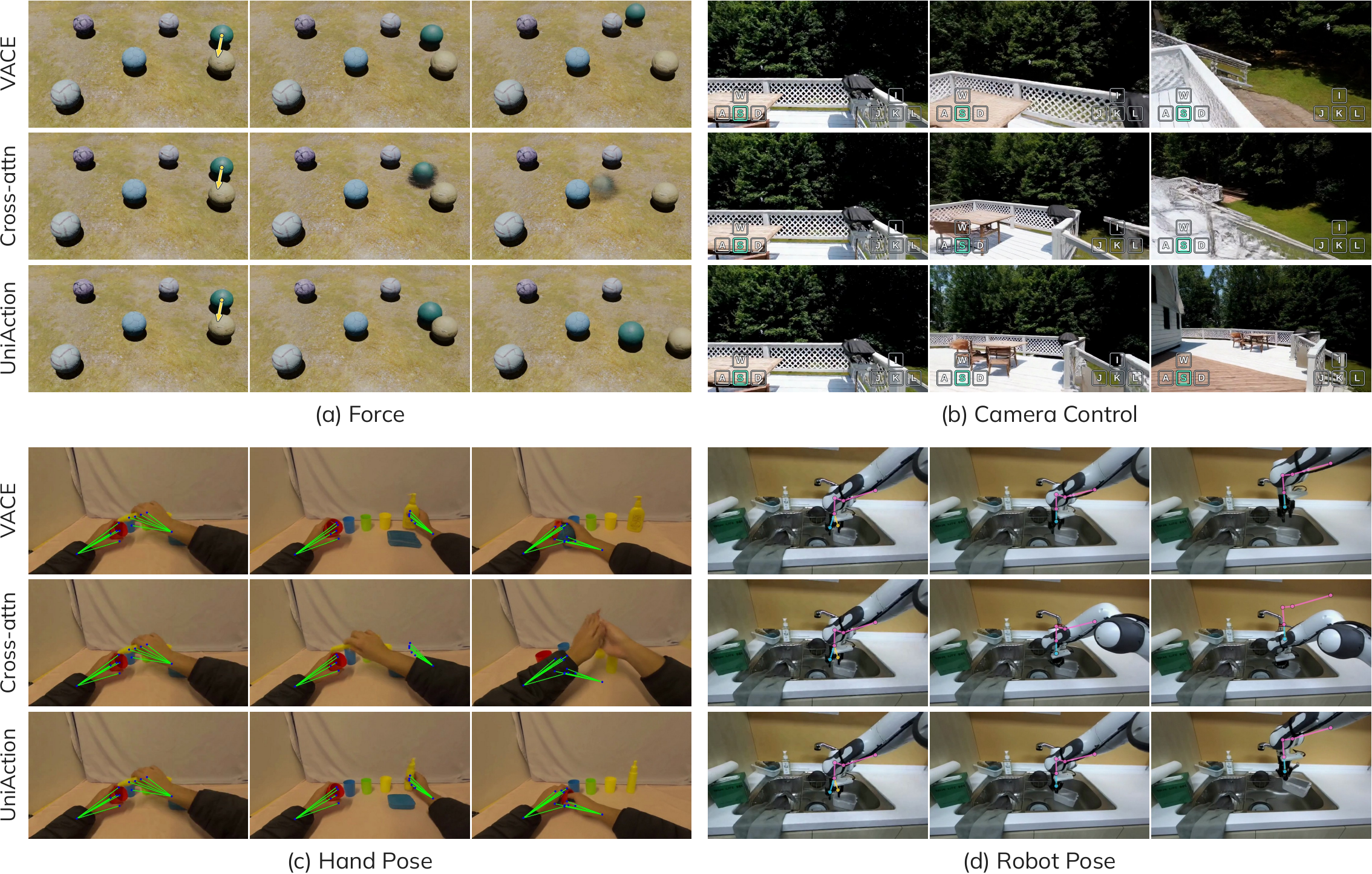}
  \caption{\textbf{Qualitative comparison across four action forms.} \ourmethod follows the input actions more faithfully than other conditioning mechanisms. Overlays indicate the input action.}
  \label{fig:qualitative_comparison}
\end{figure}

%% file: figTexts/fig_expert_comparison.tex
\begin{figure}[t]
  \centering
  \includegraphics[width=\linewidth]{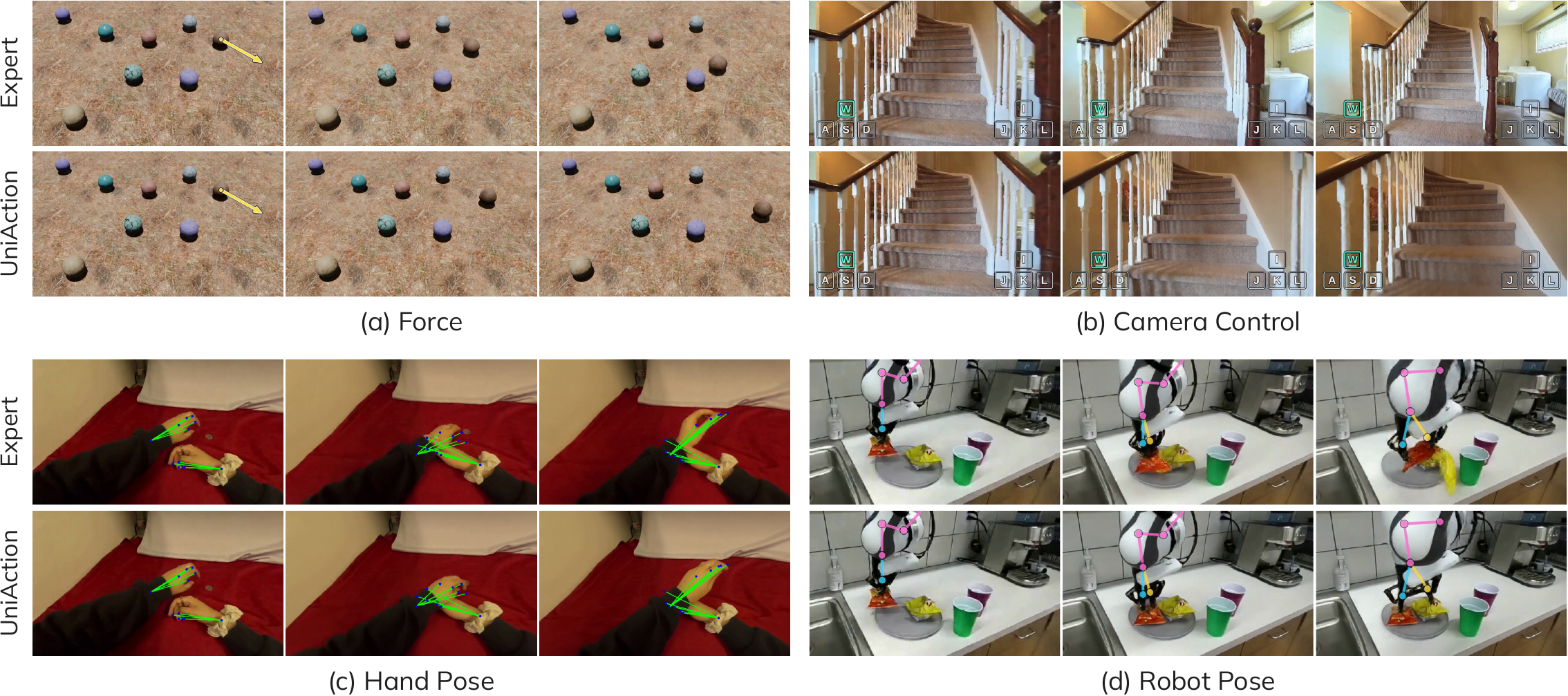}
  \caption{\textbf{Qualitative comparison with action-specific experts.} Compared with separate training, the jointly trained \ourmethod follows the input actions more faithfully and produces more plausible dynamics and visual appearance. Overlays indicate the input action.}
  \label{fig:expert_comparison}
\end{figure}

%% file: tables/joint_expert.tex
\begin{wraptable}[10]{r}{0.47\linewidth}
\vspace{-20pt}
\caption{
Jointly trained \ourmethod vs. action-specific experts. Best in \textbf{bold}.}
\label{tab:joint_learning}
\centering\small
\definecolor{uniactionrow}{gray}{0.92}
\setlength{\tabcolsep}{3.5pt}
\renewcommand{\arraystretch}{1.15}
\arrayrulecolor{black}
\newcommand{\ours}{\cellcolor{uniactionrow}}
\begin{tabular}{lcccc}
\toprule[0.8pt]
\multirow{2}{*}[-2pt]{Action} & \multicolumn{2}{c}{Action Following $\uparrow$}
& \multicolumn{2}{c}{LPIPS $\downarrow$} \\
\cmidrule(lr){2-3}\cmidrule(lr){4-5}
& Expert & Ours & Expert & Ours \\
\midrule[0.4pt]
Force & 0.5436 & \ours\textbf{0.7630} & 0.1295 & \ours\textbf{0.1163} \\
Camera & 0.5182 & \ours\textbf{0.6371} & \textbf{0.4780} & \ours 0.4863 \\
Hand & 0.3701 & \ours\textbf{0.4119} & 0.0934 & \ours\textbf{0.0910} \\
Robot & 0.6310 & \ours\textbf{0.6521} & 0.1286 & \ours\textbf{0.1164} \\
\bottomrule[0.8pt]
\end{tabular}
\vspace{-8pt}
\end{wraptable}

%% file: tables/ablation.tex
\Needspace{15\baselineskip}
\begin{wraptable}[13]{r}{0.54\linewidth}
\vspace{-22pt}
\caption{Ablations of spatiotemporal grounding and cardinality calibration. Each variant changes \emph{one} component ($A\!\to\!B$: $A$ replaced by $B$). Best in \textbf{bold}, second-best \underline{underlined}.}
\label{tab:ablation}
\centering\small
\setlength{\tabcolsep}{2.4pt}
\renewcommand{\arraystretch}{1.15}
\arrayrulecolor{black}
\begin{tabular}{@{}lcccc@{}}
\toprule[0.8pt]
\multirow{2}{*}[-2pt]{Variant} & \multicolumn{4}{c}{Action Following $\uparrow$} \\
\cmidrule(l){2-5}
& Force & Camera & Hand & Robot \\
\midrule[0.4pt]
Self-attn $\to$ Cross-attn & 0.2975 & 0.5050 & 0.0533 & 0.0733 \\
RoPE $\to$ Sinusoidal APE & 0.2084 & \underline{0.5278} & 0.0603 & 0.0755 \\
RoPE $\to$ Learned APE & 0.1485 & 0.5206 & 0.0656 & 0.0693 \\
w/o Cardinality Calib. & \underline{0.5186} & 0.4966 & \textbf{0.4303} & \underline{0.6995} \\
\midrule[0.4pt]
\rowcolor{gray!15}\textbf{UniAction (Ours)} & \textbf{0.7190} & \textbf{0.5842} & \underline{0.4266} & \textbf{0.7004} \\
\bottomrule[0.8pt]
\end{tabular}
\vspace{-10pt}
\end{wraptable}

%% file: figTexts/fig_temporal_generalization.tex
\Needspace{18\baselineskip}
\begin{wrapfigure}{r}{0.48\linewidth}
  \vspace{-10pt}
  \centering
  \setlength{\abovecaptionskip}{5pt}
  \includegraphics[width=\linewidth]{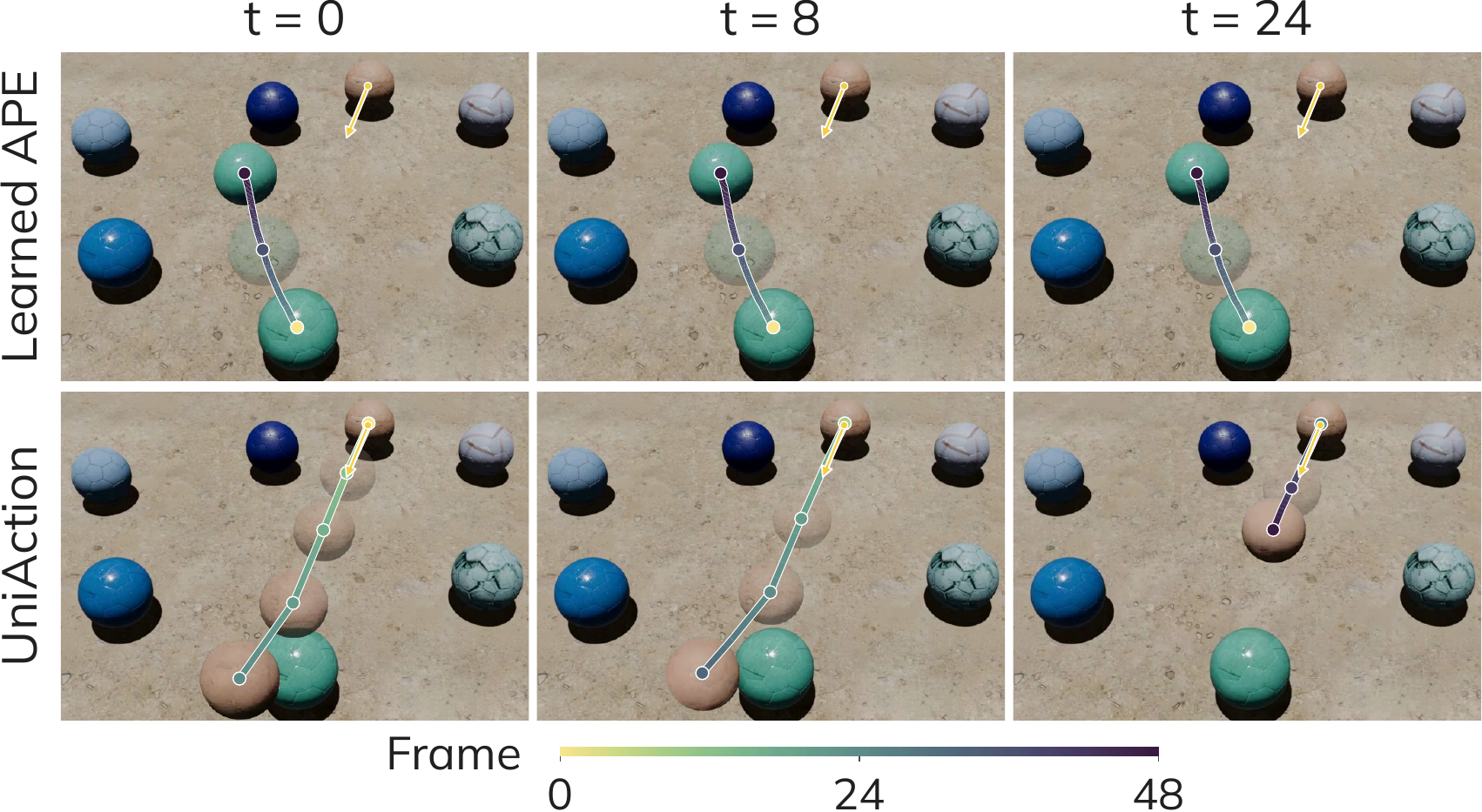}
  \caption{\textbf{Generalization.} Both models are trained only on force data, with all forces applied at $t=0$; columns show test-time forces at different $t$. Trajectories are colored by frame.}
  \label{fig:temporal_generalization}
  \vspace{-8pt}
\end{wrapfigure}

%% file: sections/05_conclusion.tex
\vspace{-3pt}
\section{Conclusion}
\vspace{-2pt}

We presented \ourmethod, a unified action-conditioning framework that represents heterogeneous actions as sparse sets of elements, each defined by the semantic identity and the spatiotemporal position. By placing them in the pretrained video model's native positional geometry, \ourmethod lets action and video tokens interact within a shared space, while cardinality-calibrated attention prevents representation granularity from biasing conditioning strength. Across four representative action types including force, camera control, hand pose, and robot pose, \ourmethod outperforms visual- and token-based conditioning alternatives, and a single jointly trained model surpasses independently trained action-specific experts, showing that heterogeneous action data can benefit one another within one world model. More broadly, \ourmethod offers an extensible interface through which new forms of intervention data can expand a world model's control capabilities.

%% file: sections/06_ai_use_statement.tex
\subsection*{AI Use Statement}

We used large language models~(LLMs) as AI tools to aid and polish the writing of this paper, including improving grammar, clarity, and conciseness of text drafted by the authors. AI tools were also used to create some decorative icons in the figures. Initial frames used as inputs for some qualitative examples in Supplementary~\ref{sec:supp_scene_generalization} and~\ref{sec:supp_compositional_generalization} were generated by GPT Image 2. In addition, an AI coding assistant was used to help produce the introductory video in the supplementary file, by writing the rendering code for its animated diagrams, layout, and on-screen captions. All experimental results shown in the video, including generated videos and quantitative numbers, are the actual outputs and reported results of our models and baselines; AI tools were not used to generate, modify, or select these results.
The authors reviewed all AI-assisted content and take full responsibility for this paper and its supplementary materials.

%% file: sections/07_ethics_statement.tex
\section*{Ethics Statement}
Our human evaluation recruited 100 participants through Prolific to rate generated videos on action following and visual quality. Participants were compensated in accordance with Prolific's payment policy and provided informed consent before participating, and no personally identifiable information was collected. All training data used in this work are drawn from publicly available datasets and are used in accordance with their licenses.

%% file: appendices/appendix.tex
\numberwithin{equation}{section}
\newcounter{suppalgorithm}

\begingroup
\setlength{\parskip}{4pt}
\setlength{\abovedisplayskip}{5pt plus 1pt minus 2pt}
\setlength{\belowdisplayskip}{5pt plus 1pt minus 2pt}
\setlength{\abovedisplayshortskip}{0pt plus 1pt}
\setlength{\belowdisplayshortskip}{3pt plus 1pt minus 1pt}
\section*{Supplementary Material}

{\raggedright
We provide an introduction video of the method in the supplementary package. We provide additional video results and comparisons on the project page: \mbox{\textbf{\url{https://uniaction.github.io/}}}\par}

\section{Semantic Representation and Extensibility}

Each semantic identity $s\in\mathcal{S}$ indexes an independently initialized action token $E[s]\in\mathbb{R}^{d}$. Under isotropic initialization, let $\bar{\mathbf e}_s = E[s]/\|E[s]\|_2$ denote its normalized representation. For two distinct identities $s\neq s'$,
\begin{equation}
    \mathbb{E}\!\left[\bar{\mathbf e}_s^\top \bar{\mathbf e}_{s'}\right]=0,
    \qquad
    \mathrm{Var}\!\left[\bar{\mathbf e}_s^\top \bar{\mathbf e}_{s'}\right]
    =\mathcal{O}\!\left(\frac{1}{d}\right).
\end{equation}
Thus, in a high-dimensional feature space, different semantic identities are initialized with low correlation and approximate orthogonality, while their correspondence to scene evolution is learned through joint action-conditioned training.

\paragraph{Extensibility.}
The semantic vocabulary can be extended by appending independently initialized tokens to the same codebook. New identities inherit the same low-correlation initialization with existing identities, allowing additional action forms to be incorporated under the same semantic representation and learned through the shared world model.

\section{Cardinality-Calibrated Attention}
\label{sec:supp_cardinality}

\subsection{Derivation}

UniAction processes video and action tokens through a shared self-attention softmax. For a video query $\mathbf{q}_v$, let the attention logits to the $j$-th video key and the $k$-th action key be
\begin{equation}
    L_{v,j}=\frac{\mathbf{q}_v^\top\mathbf{k}_{v,j}}{\sqrt{d_h}},
    \qquad
    L_{a,k}=\frac{\mathbf{q}_v^\top\mathbf{k}_{a,k}}{\sqrt{d_h}},
\end{equation}
where $d_h$ is the attention head dimension. The video and action logits are normalized by the same softmax. The total attention share assigned to the action tokens is therefore
\begin{equation}
    P_{\mathrm{action}}
    =\frac{\sum_{k=1}^{N_a}\exp(L_{a,k})}
    {\sum_{j=1}^{N_v}\exp(L_{v,j})+\sum_{k=1}^{N_a}\exp(L_{a,k})}.
\end{equation}
Define the unnormalized video and action contributions as
\begin{equation*}
    Z_v=\sum_{j=1}^{N_v}\exp(L_{v,j}),
    \qquad
    Z_a=\sum_{k=1}^{N_a}\exp(L_{a,k}),
\end{equation*}
such that $P_{\mathrm{action}}=Z_a/(Z_v+Z_a)$.

\paragraph{Cardinality bias.}
The action contribution can be rewritten as
\begin{equation}
    Z_a=N_a\cdot\operatorname{mean}_{k}\!\left[\exp(L_{a,k})\right].
\end{equation}
Here, $\operatorname{mean}_{k}[\exp(L_{a,k})]$ captures the average compatibility between the video query and the action tokens, while $N_a$ is determined by the representation granularity of the action. Consequently, actions with comparable average compatibility can receive different total attention shares solely because they are represented by different numbers of tokens.

\paragraph{Calibration.}
Cardinality calibration adds the same bias to every action logit,
\begin{equation}
    b=\log\frac{N_{\mathrm{ref}}}{N_a},
    \qquad
    L'_{a,k}=L_{a,k}+b,
\end{equation}
where $N_{\mathrm{ref}}$ is a fixed reference cardinality shared across action forms. The calibrated action contribution becomes
\begin{equation}
    \begin{aligned}
        Z'_a
        &=\sum_{k=1}^{N_a}\exp(L_{a,k}+b)
          =\frac{N_{\mathrm{ref}}}{N_a}\sum_{k=1}^{N_a}\exp(L_{a,k})\\
        &=N_{\mathrm{ref}}\cdot\operatorname{mean}_{k}\!\left[\exp(L_{a,k})\right].
    \end{aligned}
\end{equation}
\begin{samepage}
The corresponding action attention share is
\begin{equation}
    P'_{\mathrm{action}}
    =\frac{N_{\mathrm{ref}}\operatorname{mean}_{k}[\exp(L_{a,k})]}
    {Z_v+N_{\mathrm{ref}}\operatorname{mean}_{k}[\exp(L_{a,k})]}.
\end{equation}
The explicit dependence on the actual action cardinality $N_a$ is thus removed from the total action contribution, while the average action--video compatibility is preserved.
\end{samepage}

Since the same bias is applied to all action logits, $L'_{a,i}-L'_{a,j}=L_{a,i}-L_{a,j}$, so cardinality calibration preserves the relative attention relationships among action tokens while placing different action forms on a common attention scale.

\subsection{Forward, Backward, and Inference}

\paragraph{Forward computation.}
Let $L_v$ and $L_a$ denote the video and action logit matrices for the video queries. The calibrated attention weights are
\begin{equation}
    [A_v\mid A_a]=\operatorname{softmax}\!\left([L_v\mid L_a+b]\right),
    \qquad b=\log\frac{N_{\mathrm{ref}}}{N_a},
\end{equation}
and the video output is
\begin{equation}
    O_v=A_vV_v+A_aV_a.
\end{equation}
Cardinality calibration therefore enters the forward computation as a scalar offset applied to the action-logit block before the shared softmax.

\paragraph{Backward computation.}
The backward pass follows the standard softmax-attention derivatives. Let $G=\partial\mathcal{L}/\partial O_v$. For the $k$-th action token,
\begin{equation}
    \begin{aligned}
        \frac{\partial\mathcal{L}}{\partial V_{a,k}}
        &=A_{a,k}G,\\[3pt]
        \frac{\partial\mathcal{L}}{\partial L'_{a,k}}
        &=A_{a,k}G^\top\!\left(V_{a,k}-O_v\right).
    \end{aligned}
\end{equation}
The calibrated attention weight $A_{a,k}$ therefore controls both the forward contribution of each action token and the training signal propagated through the action branch.

\paragraph{Training and inference.}
The reference cardinality $N_{\mathrm{ref}}$ is fixed, and $N_a$ is determined by the current action representation. The calibration introduces no additional learnable parameters. The same calibrated attention computation is used during training and inference.

\subsection{Algorithm}

For each attention layer, cardinality calibration modifies only the action logits before the joint softmax. Given video queries $Q_v$, video keys and values $(K_v,V_v)$, and action keys and values $(K_a,V_a)$, we compute:

\noindent\begin{minipage}{\linewidth}
\refstepcounter{suppalgorithm}
\hrule height 0.8pt
\vspace{4pt}
\textbf{Algorithm \thesuppalgorithm}\quad Cardinality-calibrated attention
\par\vspace{4pt}
\hrule height 0.4pt
\vspace{4pt}
\textbf{Input:} Queries $Q_v$; keys/values $(K_v,V_v)$ and $(K_a,V_a)$; $d_h$, $N_a$, $N_{\mathrm{ref}}$.\par
\textbf{Output:} Video-token output $O_v$.
\begin{enumerate}[label=\arabic*:,leftmargin=2em,labelsep=0.7em,itemsep=3pt,topsep=4pt,parsep=0pt]
    \item $L_v \gets Q_vK_v^\top/\sqrt{d_h}$
    \item $L_a \gets Q_vK_a^\top/\sqrt{d_h}$
    \item $b \gets \log(N_{\mathrm{ref}}/N_a)$
    \item $L'_a \gets L_a+b$
    \item $[A_v\mid A_a] \gets \operatorname{softmax}([L_v\mid L'_a])$
    \item $O_v \gets A_vV_v+A_aV_a$
\end{enumerate}
\vspace{2pt}
\hrule height 0.8pt
\end{minipage}

Thus, the calibrated attention retains the original attention computation and introduces a single cardinality-dependent offset on the action-key logits.
\par
\endgroup

\input{appendices/experimental_details}
\input{appendices/efficiency}

\input{appendices/qualitative}
\input{appendices/temporal_generalization}
\input{appendices/scene_generalization}
\input{appendices/compositional_generalization}

%% file: appendices/experimental_details.tex
\clearpage
\section{Experimental Details}
\label{sec:supp_experimental_details}

\paragraph{Data and implementation.}
\looseness=-1
We select 10K action-annotated training clips from each of GoalForce~\citep{gillman2026goal}, SolarWM~\citep{huang2026solarwm}, EgoDex~\citep{hoque2026egodex}, and DROID~\citep{khazatsky2024droid}, for a total of 40K samples. Each clip is sampled to 81 frames and resized and cropped to $480\times832$ (height $\times$ width) to match the input specification of the base model. The 10K GoalForce samples consist of 5K ball, 2.5K domino, and 2.5K flower scenes.

We use the high-noise expert of Wan2.2-I2V-A14B~\citep{wan2025wan} as the base model, train it over the full noise range, and use this expert at all denoising timesteps during inference. LoRA~\citep{hu2022lora} is applied to the Q/K/V/O projections of the attention modules and the two linear layers of the FFN, with both rank and alpha set to 128. During action adaptation, only the action branch is updated, i.e., LoRA is applied only to action tokens, while the video branch remains frozen. The total training budget is two full passes over the training data. All compared methods use the same data preprocessing, per-task training data budget, and inference configuration, and are evaluated under the same task and evaluation protocol. We fix $N_{\mathrm{ref}}=784$.

Ablation studies are conducted on Wan2.2-TI2V-5B, with video clips of $480\times832$ resolution and 49 frames for both training and evaluation; all other settings follow the main experiments.

\paragraph{Test set construction.}
For each source sample, we fix the initial frame and text prompt and vary only the action condition. We use 107 source samples, none of which appear in the training set, and construct 410 test conditions per model, including both original action conditions and action-intervention variants (Table~\ref{tab:test_set}).

\begin{table}[ht]
\centering
\caption{Construction of the test set.}
\label{tab:test_set}
\small
\setlength{\tabcolsep}{4pt}
\renewcommand{\arraystretch}{1.15}
\begin{tabular}{@{}>{\raggedright\arraybackslash}p{75pt}>{\raggedright\arraybackslash}p{\dimexpr\linewidth-149pt\relax}>{\raggedleft\arraybackslash}p{58pt}@{}}
\toprule
\textbf{Action Form} & \textbf{Construction} & \textbf{\#Conditions} \\
\midrule
Force & 8 ball scenes $\times$ 5; 16 domino scenes $\times$ 2; 16 flower scenes $\times$ 2 & 104 \\
Camera Control & 13 scenes $\times$ 8 directions (\texttt{W/A/S/D/I/J/K/L}) & 104 \\
Hand Pose & 20 samples $\times$ \{original, stationary, horizontal flip, vertical flip, flip along both axes\} & 100 \\
Robot Pose & 34 samples $\times$ \{original, horizontally flipped center trajectory, reversed gripper trajectory\} & 102 \\
\midrule
\textbf{Total} & & \textbf{410} \\
\bottomrule
\end{tabular}
\end{table}

The ball variants consist of the original direction and its rotations by $\pm45^\circ$ and $\pm90^\circ$; the domino variants consist of the original direction and its $180^\circ$ reversal; the flower variants consist of the original action and a paired action that applies force in the opposite direction.

\paragraph{Action Following scores.}
Let $C(x)=\min(1,\max(0,x))$. Each task uses the following fixed scoring rule.
\begin{itemize}[leftmargin=*,topsep=3pt,itemsep=4pt,parsep=0pt]
\item \textbf{Force.} For ball and domino scenes, we estimate the initial motion response of the generated video with DIS optical flow~\citep{kroeger2016dis}. For flower scenes, since the bending and swaying direction of the flower does not necessarily coincide with the direction of the applied force, we extract DIS optical flow features at multiple locations and estimate the force direction with a ridge regression mapping. The mapping is fitted only on GoalForce training data and kept fixed throughout the evaluation of all models. Let $\hat{\mathbf f}$ and $\mathbf f$ denote the estimated and target force directions, respectively. The score is
\[
s_{\mathrm F}=C\!\left(\frac{\hat{\mathbf f}^{\top}\mathbf f}{\|\hat{\mathbf f}\|\,\|\mathbf f\|}\right).
\]
Samples that do not produce a valid motion response receive a score of 0.

\item \textbf{Hand Pose and Robot Pose.} We use CoTracker3~\citep{karaev2025cotracker3} to track the locations corresponding to the action annotations, and compare the actual displacement $\Delta\mathbf p$ relative to the initial position with the target displacement $\Delta\mathbf c$. The score is obtained by normalizing the error between the actual and target displacements, so that results more consistent with the target motion receive higher scores. For the stationary condition, the score is based on the residual motion relative to the initial position.

\item \textbf{Camera Control.} We estimate the camera trajectory with ViPE~\citep{huang2025vipe}, taking the first frame as the origin. To account for camera drift inherent to the model and scene, we use the no-action generation of the same scene and model as a reference and subtract this baseline from the predicted response. We measure the cosine similarity
between the residual motion vector and the commanded direction, separately for translation and rotation, normalize it to $[0,1]$, and report the average score.

\end{itemize}
All Action Following scores lie in $[0,1]$, with higher scores indicating better action following. We first average the different action variants of each source sample with equal weights and then aggregate over all source samples. For Force, the ball, domino, and flower subcategories are further averaged with equal weights.

\Needspace{4\baselineskip}
\paragraph{LPIPS.}
We compute AlexNet-based LPIPS~\citep{zhang2018lpips} only under the original action condition, comparing generated frames with the corresponding ground-truth frames at $480\times832$ resolution. Since the first frame is given as model input, we exclude it, compute LPIPS on each of the remaining 80 frames, and average the results.

\Needspace{10\baselineskip}
\paragraph{Human evaluation.}
A total of 100 participants completed the questionnaire. Each participant was randomly assigned 8 groups of videos (2 per action form), where each group contains one video generated by each of UniAction, VACE, and Cross-Attention, resulting in 24 videos per participant. The three videos in each group are displayed side by side in a randomized order. Referring to the action markers overlaid on the videos, participants rate each video on a 1--5 scale along the following two dimensions:
\begin{itemize}[leftmargin=*,topsep=3pt,itemsep=4pt,parsep=0pt]
\item \textbf{Q1 (Action Following):}\\*
``Does the video do what the action says? (Compare with the other videos.)''
\item \textbf{Q2 (Visual Quality):}\\*
``How good is the video quality, including clarity, stability, and plausible physical interactions? (Compare with the other videos.)''
\end{itemize}

%% file: appendices/efficiency.tex
\Needspace{21\baselineskip}
\section{Efficiency Analysis}

We measure action-conditioning overhead for one denoising forward pass on a single GH200, relative to the backbone without action input (1678.4 TFLOPs and 3.06\,s). Table~\ref{tab:efficiency} reports trainable parameters, additional TFLOPs, and relative latency increases.

VACE runs a separate conditioning branch at full video resolution, adding approximately 21\% to both computation and latency regardless of the action form. Cross Attention also has nearly fixed overhead. UniAction only appends action tokens, so its overhead varies with the number of action tokens. It incurs the lowest additional latency across all four action forms.

\input{tables/efficiency}

%% file: tables/efficiency.tex
\begin{table}[h]
\centering
\caption{Additional cost of action conditioning per denoising forward pass, relative to the backbone without action input.}
\label{tab:efficiency}
\small
\setlength{\tabcolsep}{4pt}
\renewcommand{\arraystretch}{1.15}
\begin{tabular}{lc*{8}{c}}
\toprule
\multirow{2}{*}[-2pt]{Method} & \multirow{2}{*}[-2pt]{Params}
& \multicolumn{4}{c}{$\Delta$TFLOPs} & \multicolumn{4}{c}{$\Delta$Latency (\%)} \\
\cmidrule(lr){3-6}\cmidrule(lr){7-10}
& & Force & Camera & Hand & Robot & Force & Camera & Hand & Robot \\
\midrule
VACE & 3.1B & 351.2 & 351.2 & 351.2 & 351.2 & 21.3 & 21.0 & 21.2 & 21.1 \\
Cross Attention & 640M & 33.5 & 35.9 & 36.0 & 36.0 & 8.8 & 8.6 & 9.0 & 9.0 \\
\rowcolor{gray!15}\textbf{UniAction} & 404M & 0.1 & 5.9 & 72.9 & 37.2 & 3.6 & 3.0 & 6.6 & 4.9 \\
\bottomrule
\end{tabular}
\end{table}

%% file: appendices/qualitative.tex
\clearpage
\section{Additional Qualitative Comparisons}
\label{sec:supp_qualitative}

Figures~\ref{fig:supp_qualitative_1}--\ref{fig:supp_qualitative_3} provide 13 additional comparisons across the four action forms. Each example shows six frames per method, ordered from left to right. Arrows, key indicators, and pose overlays visualize the input action conditions rather than generated content.

\begingroup
\setlength{\abovecaptionskip}{4pt}
\setlength{\belowcaptionskip}{0pt}
\setlength{\floatsep}{6pt}
\setlength{\textfloatsep}{6pt}
\setlength{\intextsep}{6pt}
\makeatletter
\setlength{\@fptop}{0pt}
\setlength{\@fpbot}{0pt plus 1fil}
\makeatother

\begin{figure}[!ht]
  \centering
  \includegraphics[width=\linewidth]{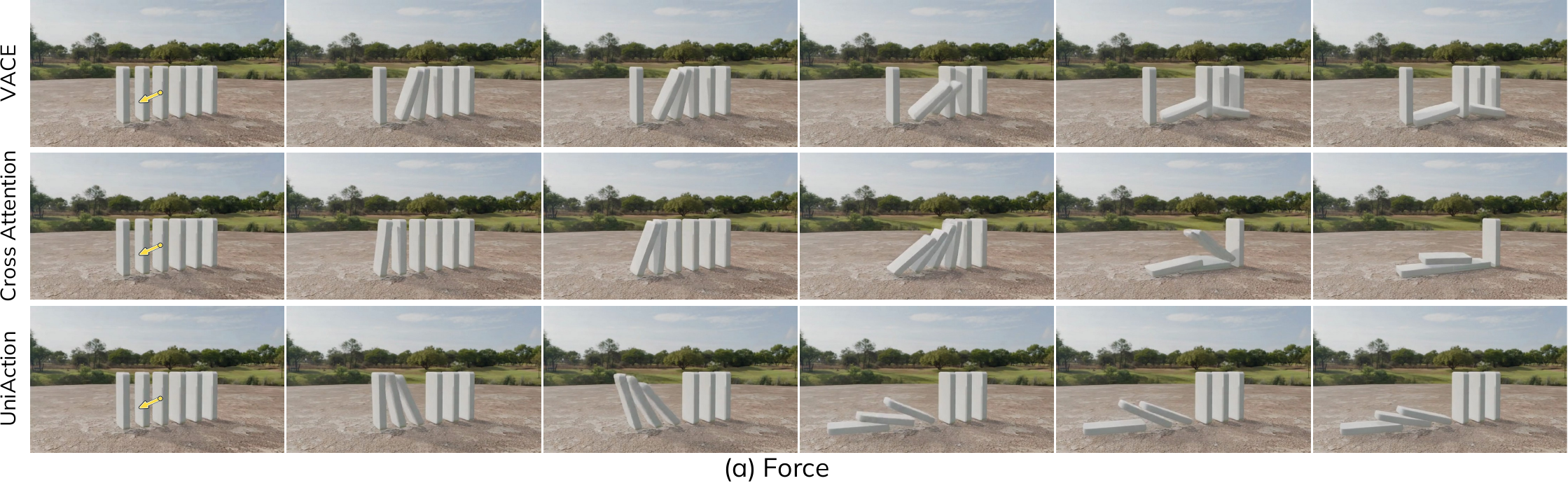}\par\vspace{2pt}
  \includegraphics[width=\linewidth]{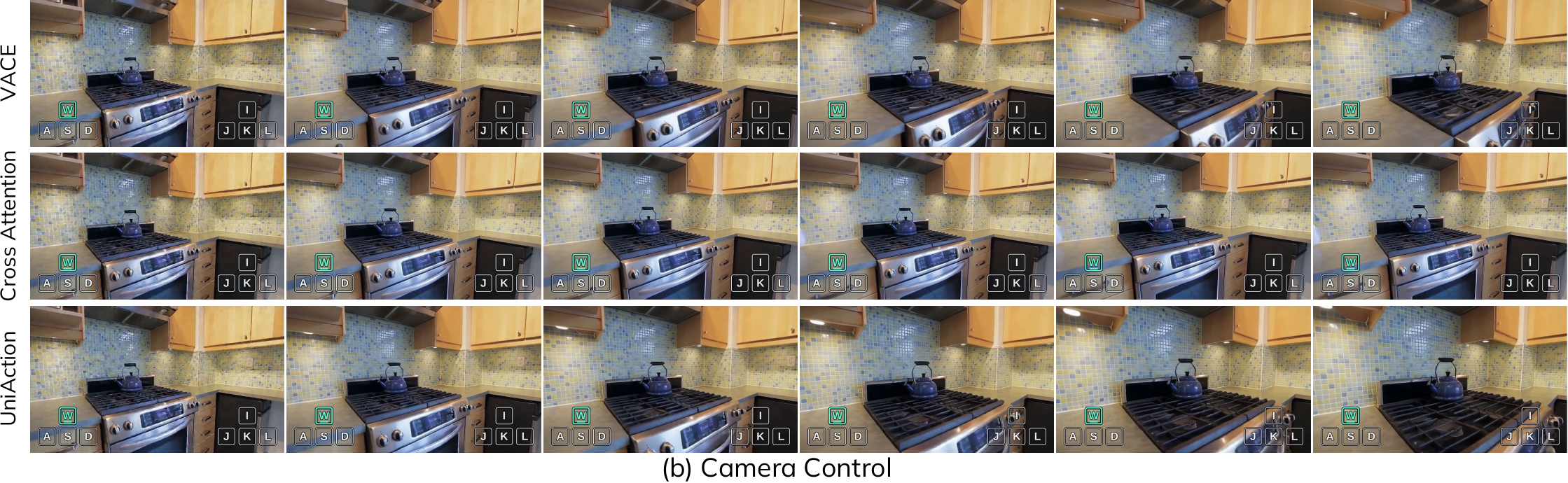}\par\vspace{2pt}
  \includegraphics[width=\linewidth]{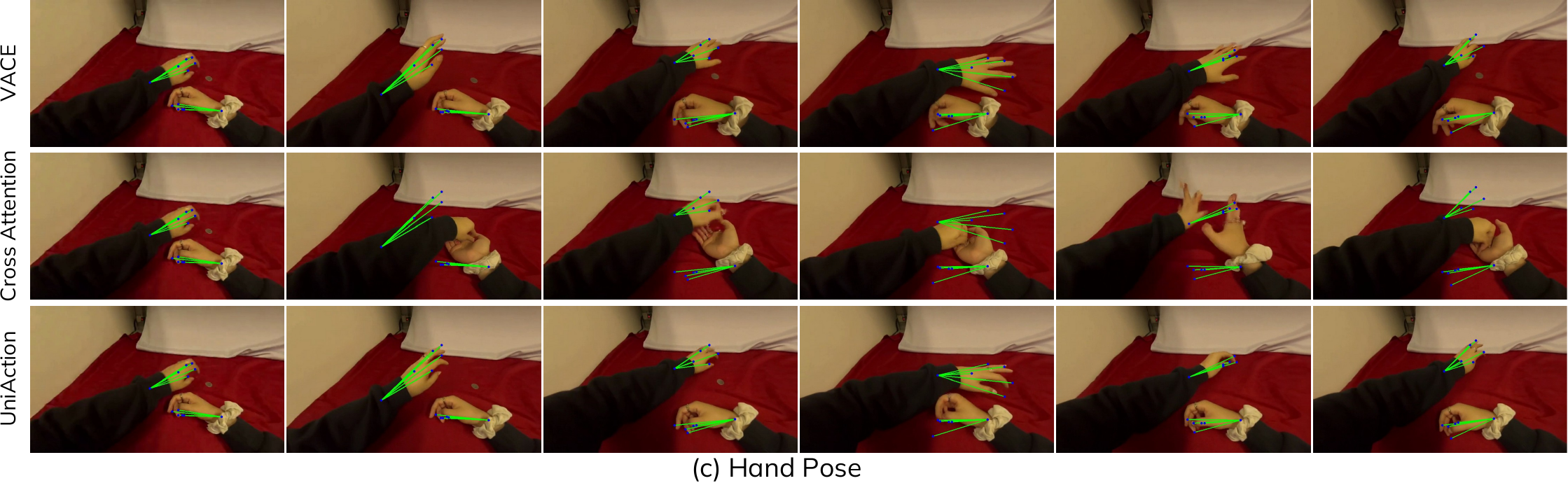}\par\vspace{2pt}
  \includegraphics[width=\linewidth]{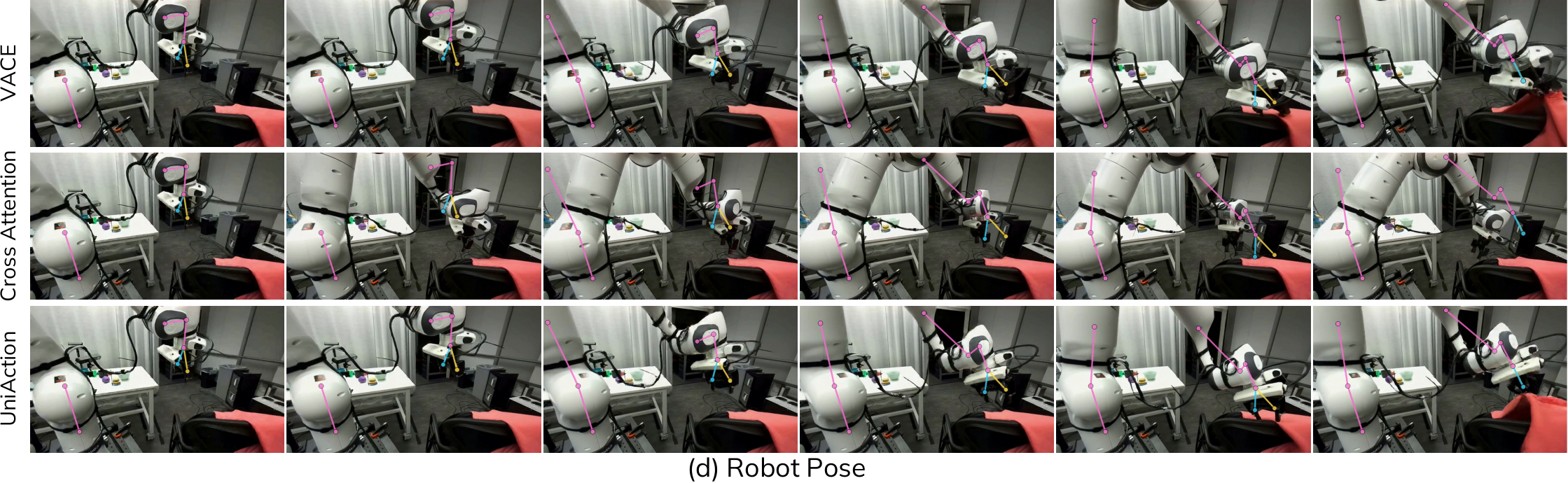}\par
  \caption{\textbf{Additional qualitative comparisons.} The camera input in (b) is \texttt{W} (forward camera motion). All overlays indicate input conditions.}
  \label{fig:supp_qualitative_1}
\end{figure}

\clearpage
\begin{figure}[t]
  \centering
  \includegraphics[width=\linewidth]{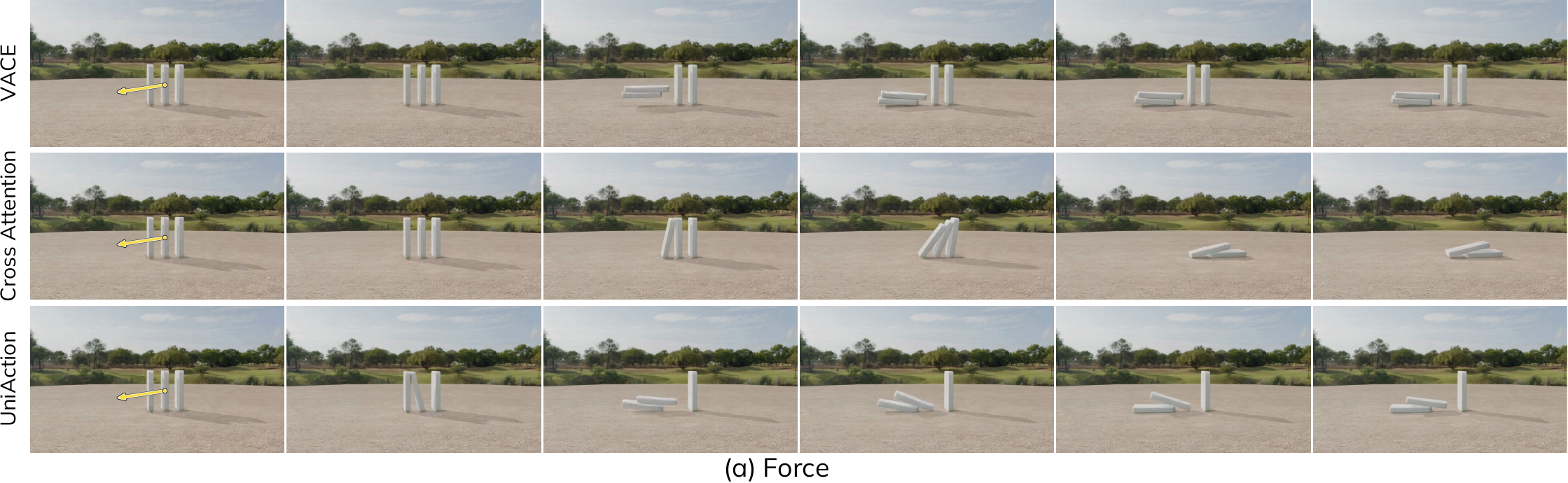}\par\vspace{1.5pt}
  \includegraphics[width=\linewidth]{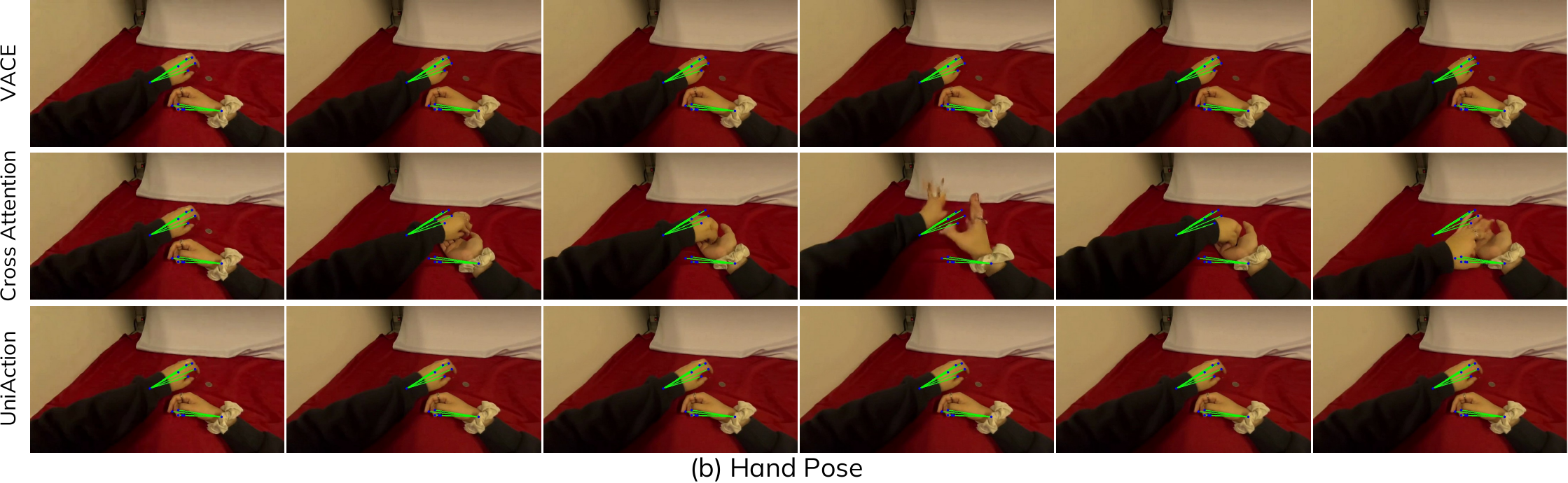}\par\vspace{1.5pt}
  \includegraphics[width=\linewidth]{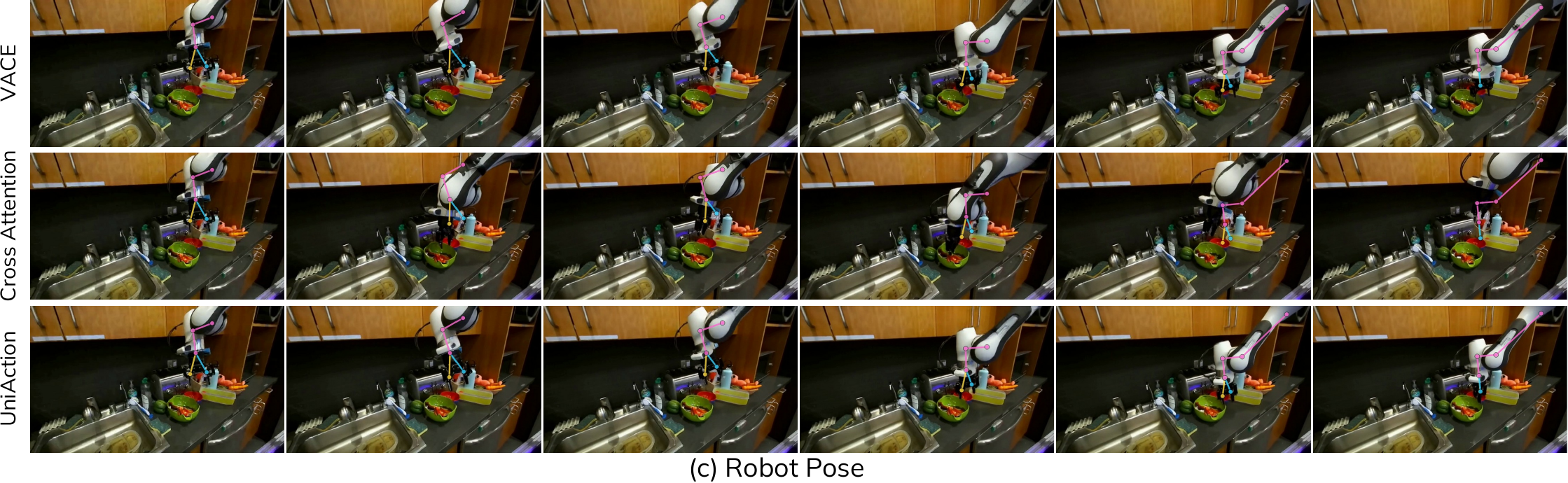}\par\vspace{1.5pt}
  \includegraphics[width=\linewidth]{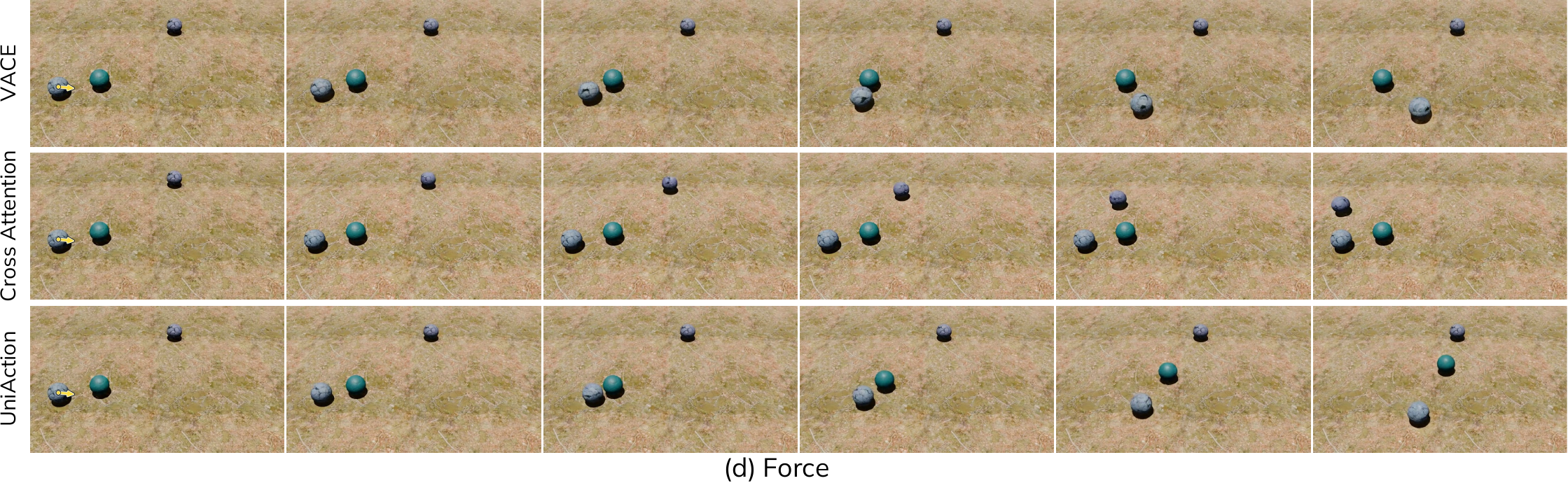}\par\vspace{1.5pt}
  \includegraphics[width=\linewidth]{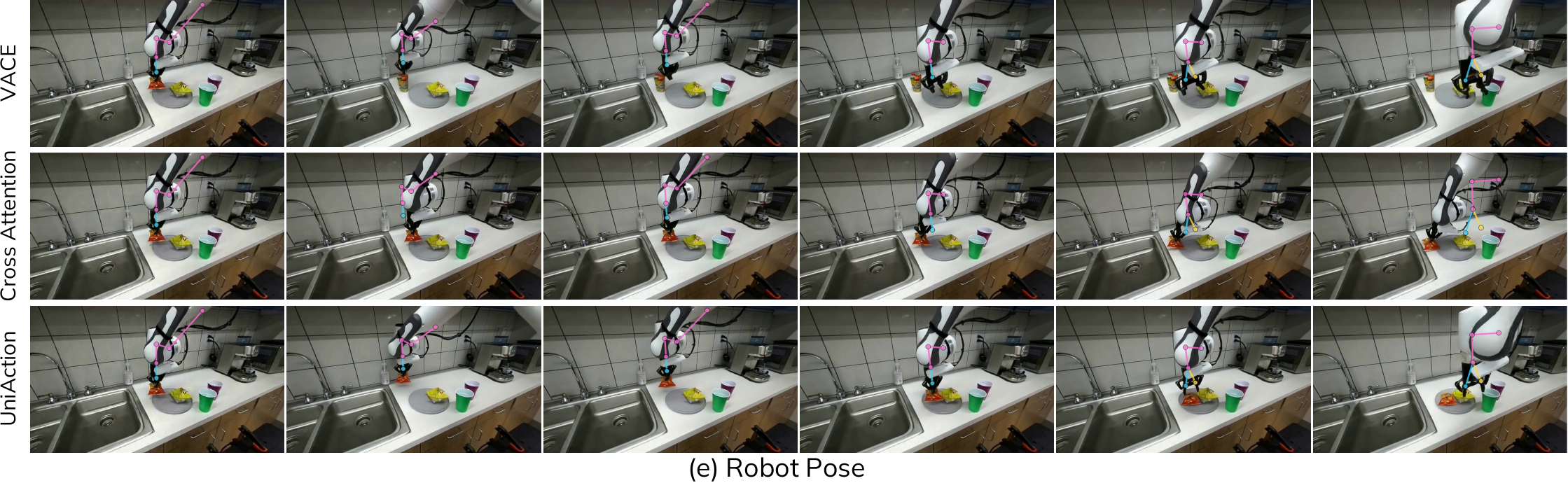}\par
  \caption{\textbf{Additional qualitative comparisons.} Overlays indicate input conditions.}
\end{figure}

\clearpage
\begin{figure}[t]
  \centering
  \includegraphics[width=\linewidth]{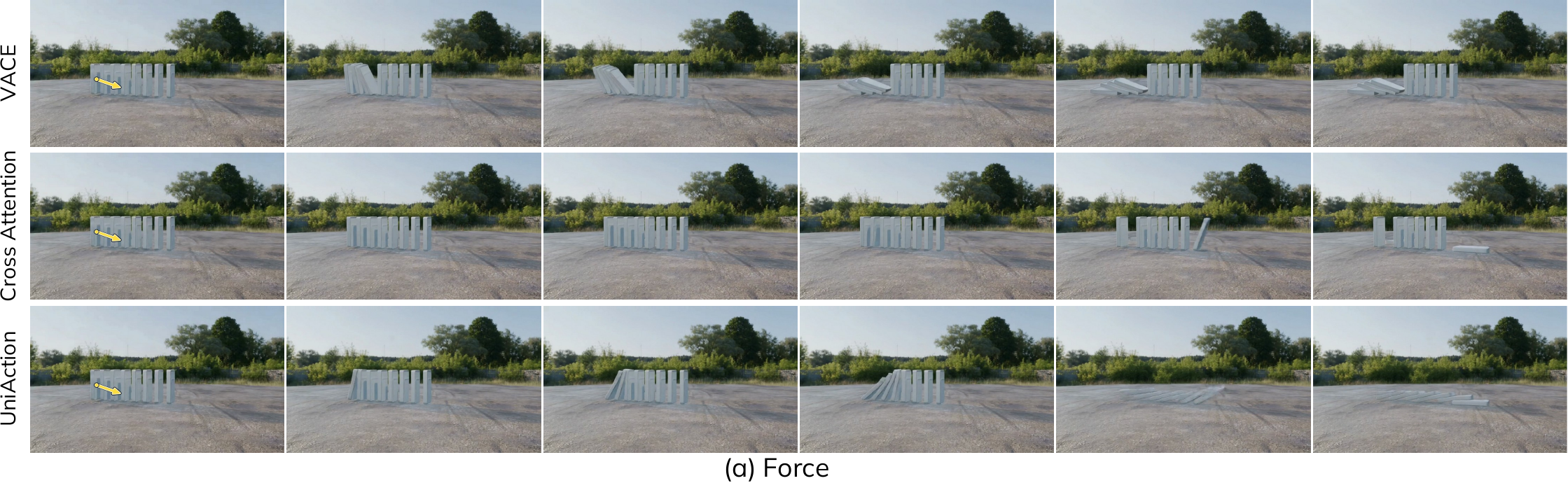}\par\vspace{2pt}
  \includegraphics[width=\linewidth]{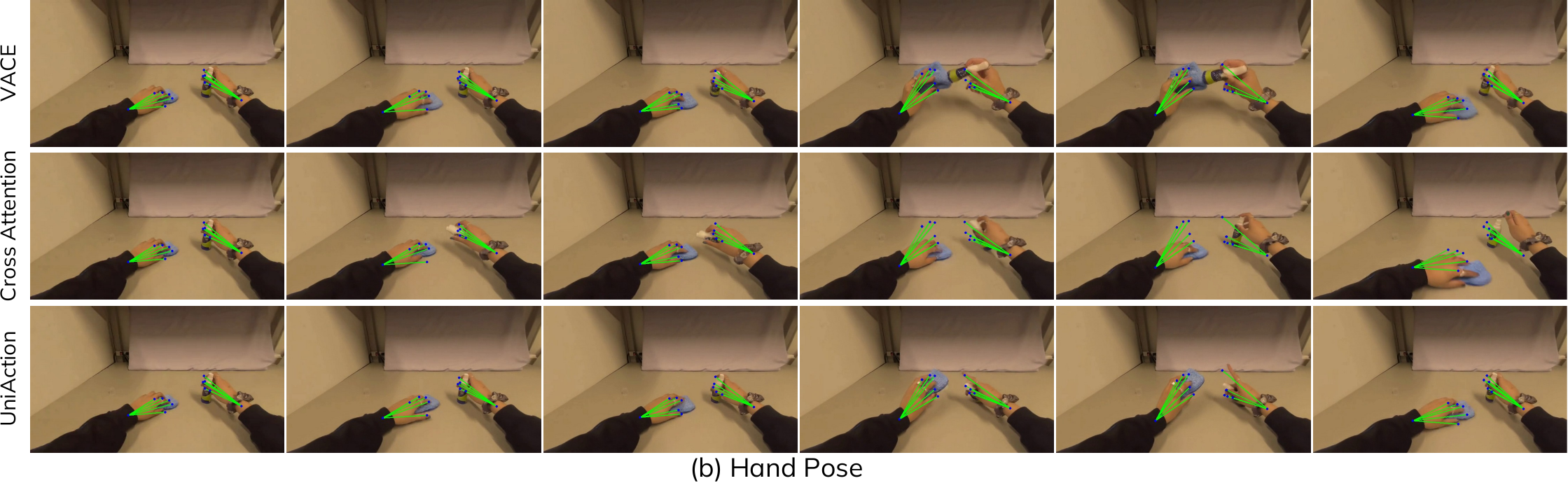}\par\vspace{2pt}
  \includegraphics[width=\linewidth]{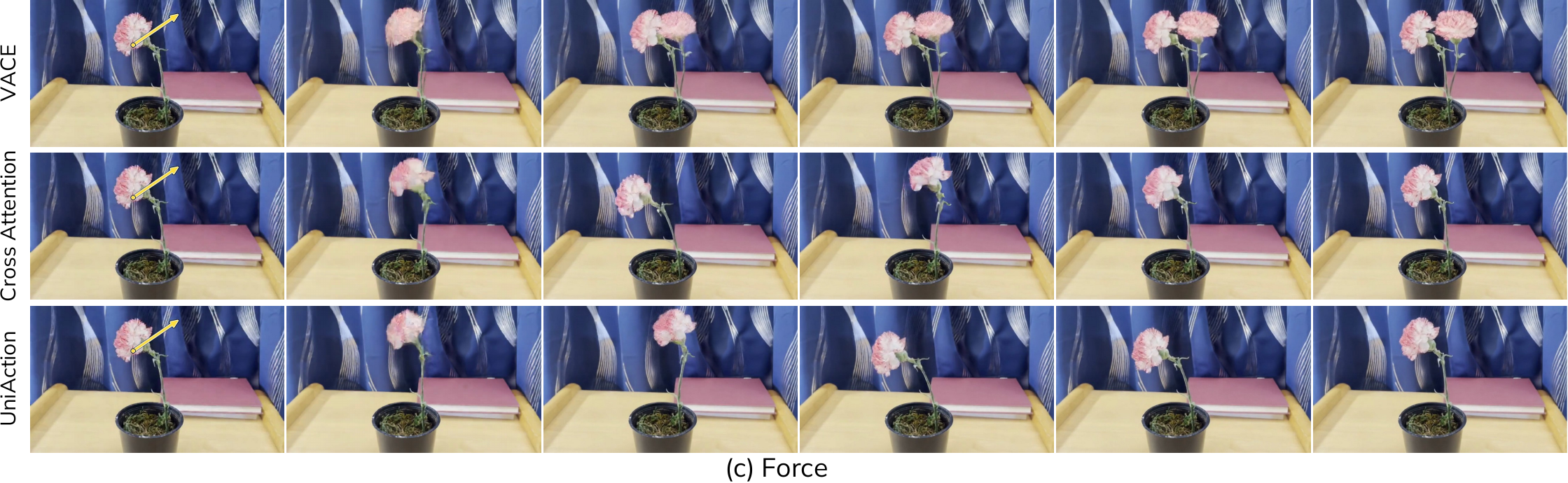}\par\vspace{2pt}
  \includegraphics[width=\linewidth]{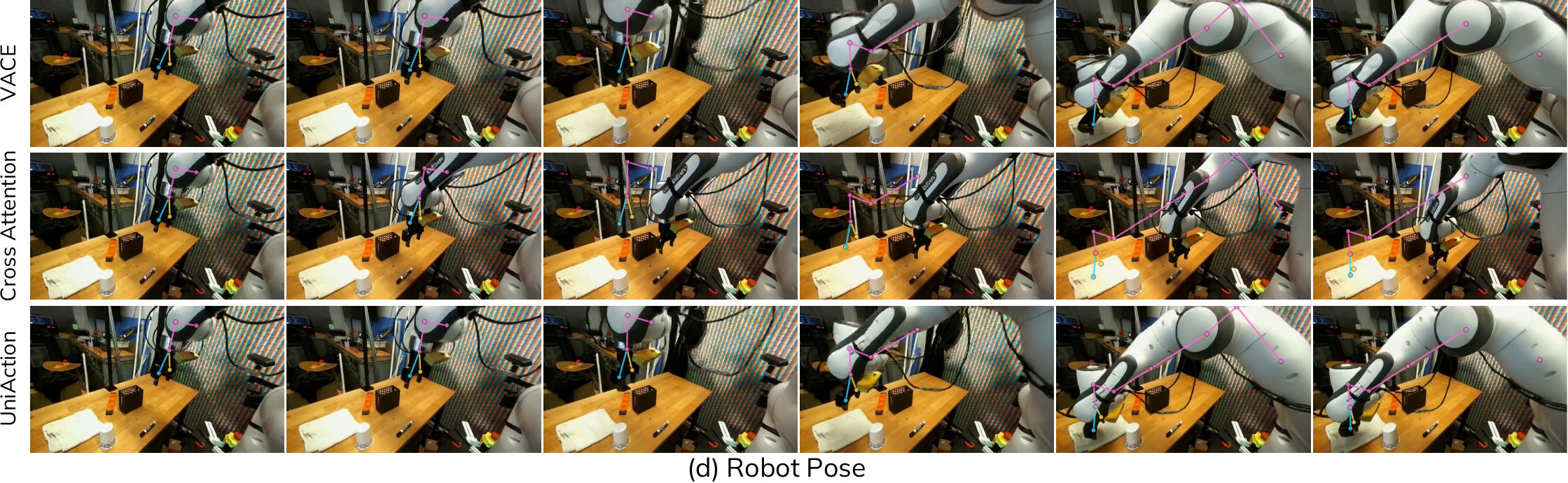}\par
  \caption{\textbf{Additional qualitative comparisons.} Overlays indicate input conditions.}
  \label{fig:supp_qualitative_3}
\end{figure}
\clearpage
\endgroup

%% file: appendices/temporal_generalization.tex
\clearpage
\section{Additional Temporal Generalization Results}

Following the setup in Sec.~\ref{sec:ablation_grounding} and Figure~\ref{fig:temporal_generalization}, we train UniAction only on force data, with all forces applied at $t=0$. Figures~\ref{fig:supp_temporal_generalization_domino} and~\ref{fig:supp_temporal_generalization_balls} show that delaying force application to $t=8$ or $t=24$ delays object motion accordingly, demonstrating generalization to unseen intervention times. Yellow arrows indicate the input force.

\begingroup
\setlength{\intextsep}{6pt}
\setlength{\abovecaptionskip}{4pt}
\setlength{\belowcaptionskip}{0pt}
\begin{figure}[!ht]
  \centering
  \includegraphics[width=\linewidth]{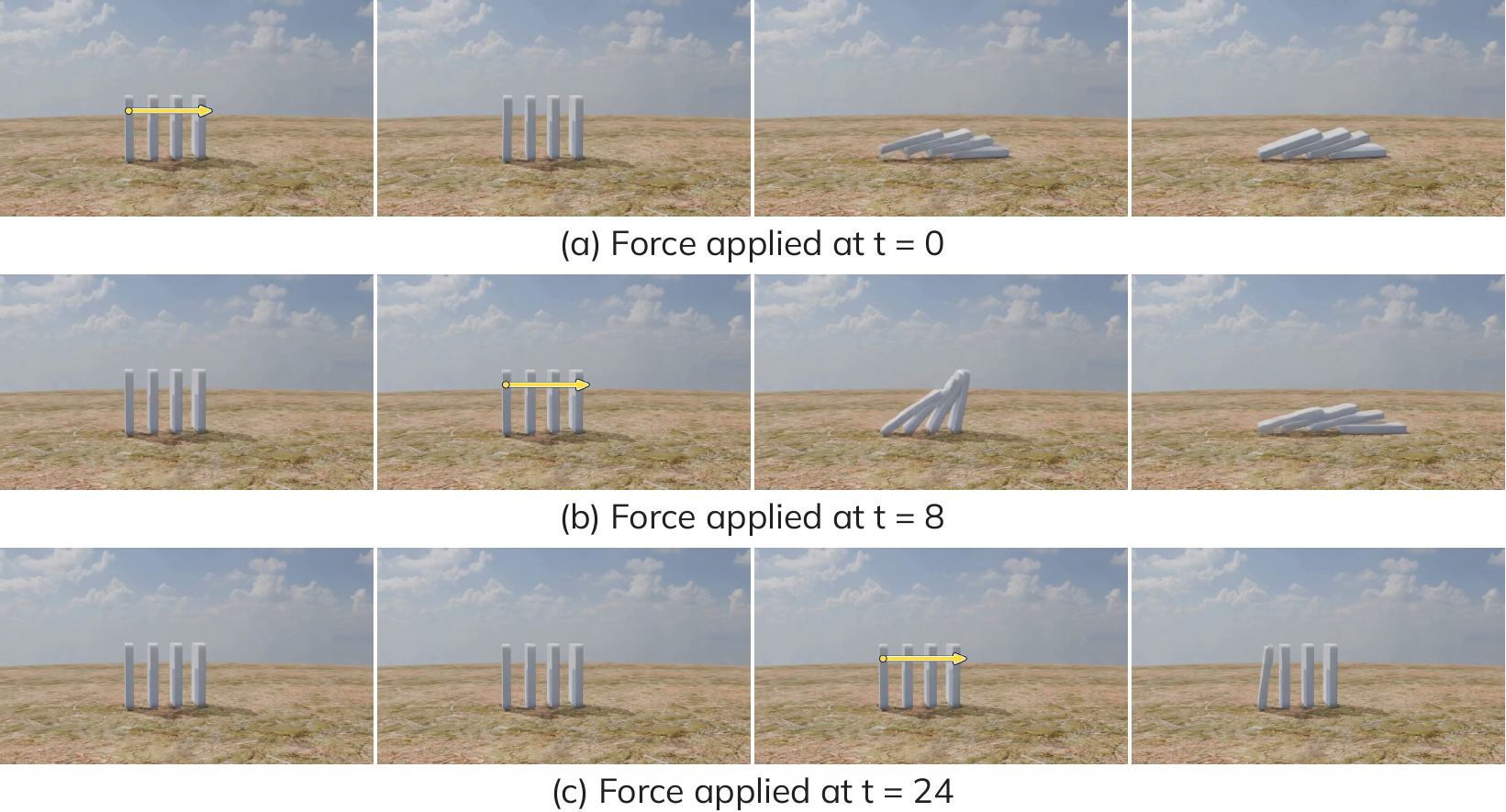}
  \caption{\textbf{Temporal generalization on dominoes.}}
  \label{fig:supp_temporal_generalization_domino}
\end{figure}

\begin{figure}[!ht]
  \centering
  \includegraphics[width=\linewidth]{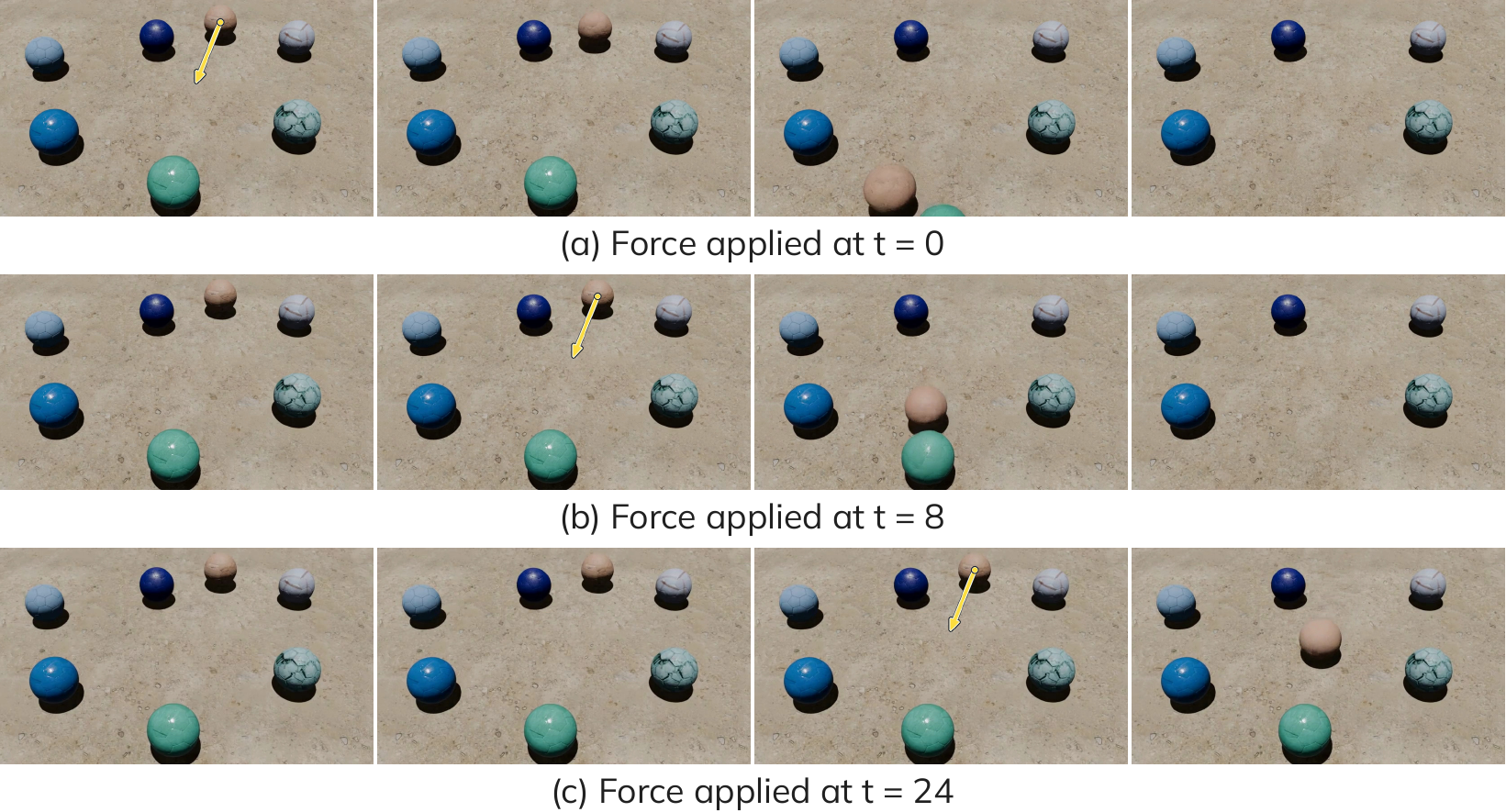}
  \caption{\textbf{Temporal generalization on balls.}}
  \label{fig:supp_temporal_generalization_balls}
\end{figure}
\endgroup

%% file: appendices/scene_generalization.tex
\clearpage
\section{Scene Generalization}
\label{sec:supp_scene_generalization}

Figure~\ref{fig:supp_scene_generalization} shows 24 generations from the jointly trained UniAction. Except for the second row in the left column of (c) and the upper-left example in (d), whose initial frames come from the test set, all initial frames were generated by GPT Image 2 and were unseen during training. Each strip shows the initial frame and two later frames; paired left and right strips use matching frame indices.

\begingroup
\setlength{\intextsep}{4pt}
\setlength{\abovecaptionskip}{4pt}
\setlength{\belowcaptionskip}{0pt}
\begin{figure}[!ht]
  \centering
  \includegraphics[width=\linewidth,height=0.76\textheight,keepaspectratio]{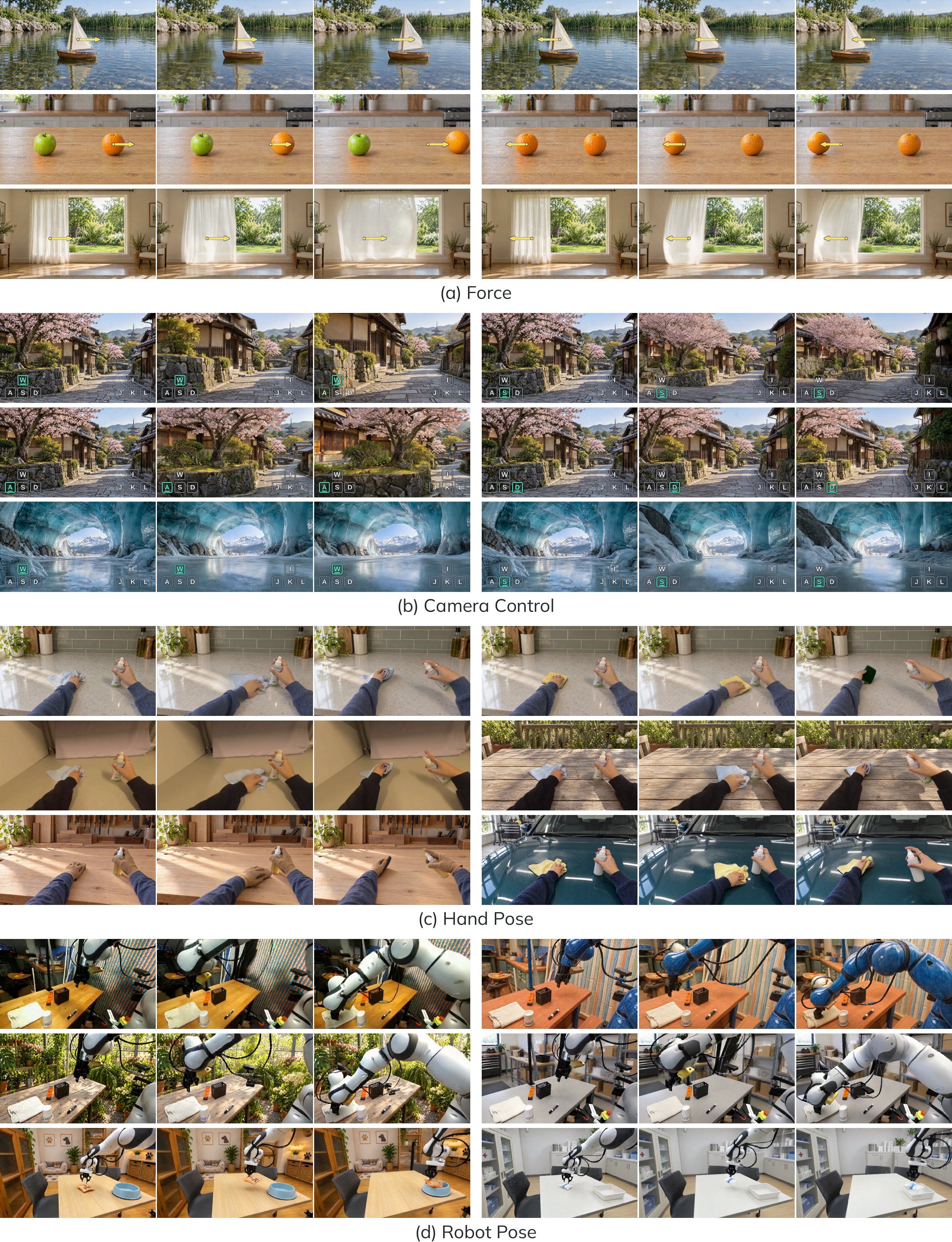}
  \caption{\textbf{Scene and appearance generalization across four action forms.} All six variants in (c) share the same hand-pose input. Arrows and highlighted keys indicate the input actions.}
  \label{fig:supp_scene_generalization}
\end{figure}
\endgroup

%% file: appendices/compositional_generalization.tex
\clearpage
\section{Compositional Action Generalization}
\label{sec:supp_compositional_generalization}

Figure~\ref{fig:supp_compositional_generalization} shows three long videos generated by the jointly trained UniAction. Each training sample contains only one action form. Since all actions share the same representation and conditioning pathway, tokens from different action forms can be directly concatenated into a single sequence at inference time without additional training. These results combine force, camera control, hand pose, and robot pose, with both simultaneous actions and transitions between actions over time. The initial frames were generated by GPT Image 2.

\begingroup
\setlength{\intextsep}{4pt}
\setlength{\abovecaptionskip}{4pt}
\setlength{\belowcaptionskip}{0pt}
\begin{figure}[!ht]
  \centering
  \includegraphics[width=\linewidth]{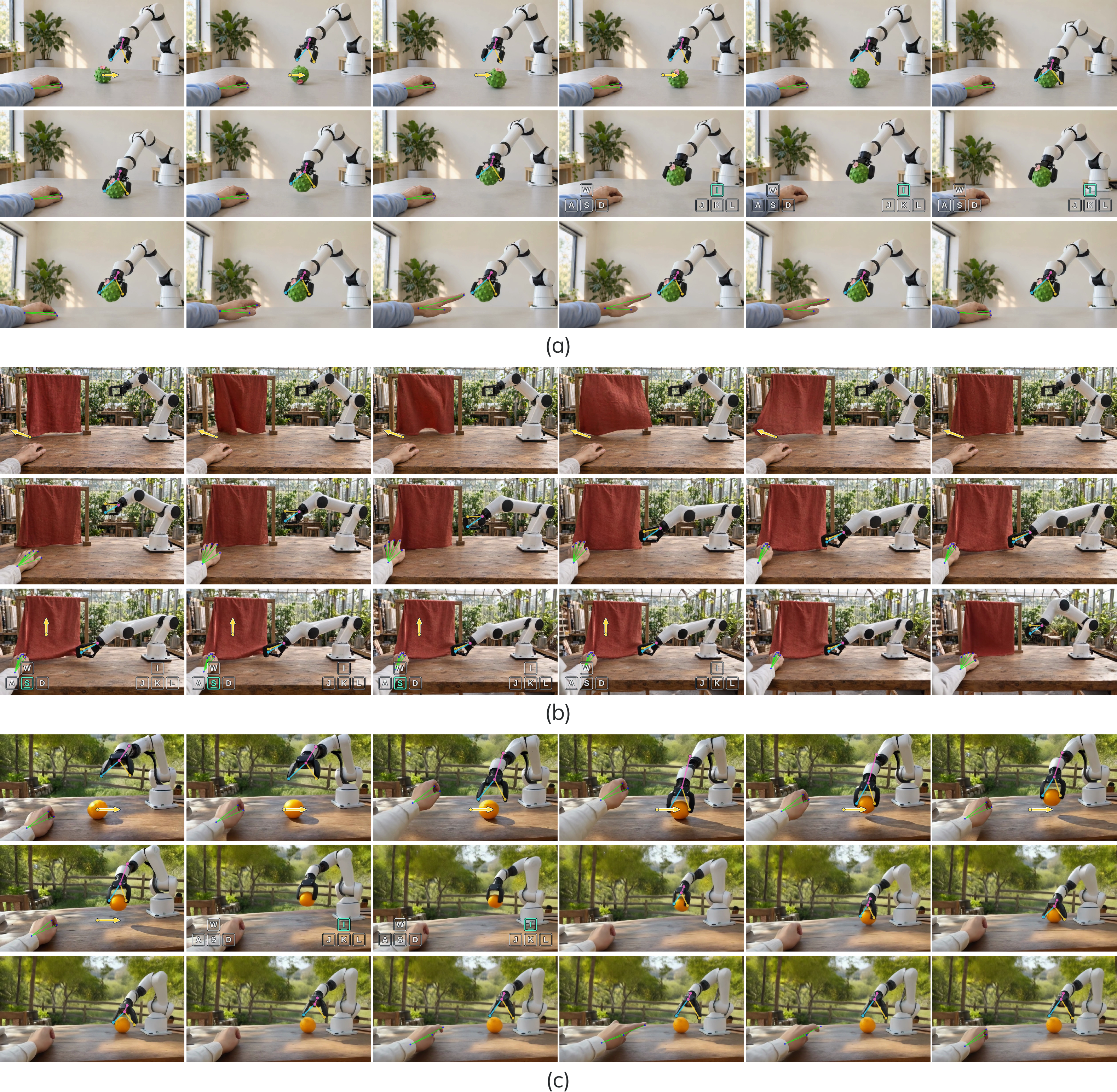}
  \caption{\textbf{Generalization to unseen action compositions.} Each block (a)--(c) shows one video in three rows of six selected frames, read from left to right and then from top to bottom. Arrows, pose overlays, and highlighted keys indicate the input actions.}
  \label{fig:supp_compositional_generalization}
\end{figure}
\endgroup